\PassOptionsToPackage{unicode}{hyperref}
\PassOptionsToPackage{hyphens}{url}
\documentclass[
]{article}
\usepackage{amsmath,amssymb}
\usepackage{iftex}
\ifPDFTeX
  \usepackage[T1]{fontenc}
  \usepackage[utf8]{inputenc}
  \usepackage{textcomp} 
\else 
  \usepackage{unicode-math} 
  \defaultfontfeatures{Scale=MatchLowercase}
  \defaultfontfeatures[\rmfamily]{Ligatures=TeX,Scale=1}
\fi
\usepackage{lmodern}
\ifPDFTeX\else
\fi
\IfFileExists{upquote.sty}{\usepackage{upquote}}{}
\IfFileExists{microtype.sty}{
  \usepackage[]{microtype}
  \UseMicrotypeSet[protrusion]{basicmath} 
}{}
\makeatletter
\@ifundefined{KOMAClassName}{
  \IfFileExists{parskip.sty}{%
    \usepackage{parskip}
  }{
    \setlength{\parindent}{0pt}
    \setlength{\parskip}{6pt plus 2pt minus 1pt}}
}{
  \KOMAoptions{parskip=half}}
\makeatother
\usepackage{xcolor}
\usepackage[margin=1in]{geometry}
\usepackage{listings}
\newcommand{\passthrough}[1]{#1}
\usepackage{longtable,booktabs,array}
\usepackage{calc} 
\usepackage{etoolbox}
\makeatletter
\patchcmd\longtable{\par}{\if@noskipsec\mbox{}\fi\par}{}{}
\makeatother
\IfFileExists{footnotehyper.sty}{\usepackage{footnotehyper}}{\usepackage{footnote}}
\makesavenoteenv{longtable}
\usepackage{graphicx}
\makeatletter
\def\maxwidth{\ifdim\Gin@nat@width>\linewidth\linewidth\else\Gin@nat@width\fi}
\def\maxheight{\ifdim\Gin@nat@height>\textheight\textheight\else\Gin@nat@height\fi}
\makeatother
\setkeys{Gin}{width=\maxwidth,height=\maxheight,keepaspectratio}
\makeatletter
\def\fps@figure{htbp}
\makeatother
\providecommand{\tightlist}{%
  \setlength{\itemsep}{0pt}\setlength{\parskip}{0pt}}
\AtBeginEnvironment{longtable}{\footnotesize}
\usepackage{listings}
\usepackage{fvextra}
\fvset{breaklines=true,breakanywhere=true,breaksymbolleft={},fontsize=\small}

\IfFontExistsTF{Noto Sans Mono CJK KR}{%
  \setmonofont{Noto Sans Mono CJK KR}[Scale=0.92]%
}{}
\ifLuaTeX
  \usepackage{selnolig}  
\fi
\IfFileExists{bookmark.sty}{\usepackage{bookmark}}{\usepackage{hyperref}}
\IfFileExists{xurl.sty}{\usepackage{xurl}}{} 
\hypersetup{
  hidelinks,
  pdfcreator={LaTeX via pandoc}}

\author{}
\date{}

\begin{document}

\hypertarget{measuring-and-improving-behavioral-consistency-in-large-language-models-through-factheuristicemotion-state-enforcement}{%
\section{Measuring and Improving Behavioral Consistency in Large
Language Models through Fact--Heuristic--Emotion State
Enforcement}\label{measuring-and-improving-behavioral-consistency-in-large-language-models-through-factheuristicemotion-state-enforcement}}

\textbf{Authors:}\\
Gi-Hun Lee\(^{1,\ast}\), Joong Yull Park\(^{2,\ast}\)

\(^{1}\) Independent Researcher\\
\(^{2}\) School of Mechanical Engineering, Chung-Ang University, Seoul,
Korea\\
\(^{\ast}\) Corresponding authors: leescoppermine@gmail.com (G.-H. Lee),
jrpark@cau.ac.kr (J. Y. Park)

\textbf{Preprint:} arXiv (cs.AI, cs.HC)\\
\textbf{Version:} arXiv v1.0\\
\textbf{Date:} 2026-06-05

\begin{center}\rule{0.5\linewidth}{0.5pt}\end{center}

\hypertarget{abstract}{%
\subsection{Abstract}\label{abstract}}

Large language models (LLMs) can give different answers to the same
decision problem across repeated runs. They can also change their
decisions when their own previous answer is reintroduced as context.
This paper asks whether that instability can be measured and partially
reduced without changing model weights.

We test a lightweight intervention called the \textbf{Cognitive Kernel
Model (CKM)}. Before making a decision, the model must separate the
input into three epistemic roles: \textbf{Fact} (what is given or
verifiable), \textbf{Heuristic} (what is inferred or assumed), and
\textbf{Emotion} (what carries evaluative or priority signal). In
practical terms, CKM is a prompt-level state-enforcement layer. It does
not make the model more capable; it forces the model to track what kind
of information it is using before it acts.

Formally, CKM maintains a structured state

\[S_t = \{F_t, H_t, E_t\}\]

and updates it through a state transition function:

\[S_{t+1} = \delta(S_t, \Omega_{x,t}, \Omega^{\text{Engine}}_t).\]

We evaluated CKM on Korean-language student decision scenarios involving
ambiguity resolution, ethical conflict, resource allocation, and error
handling. The evaluation includes four core experiments, a 4-arm
ablation, a 5-arm sham-restriction ablation, and a temperature
robustness probe. The dataset spans 26 unique LLM models from four
vendors (OpenAI, Anthropic, Google, xAI), two model generations, and
37,403 total observations (35,475 primary + 1,928 case-study
observations).

Six findings emerged.

\begin{enumerate}
\def\labelenumi{\arabic{enumi}.}
\item
  \textbf{CKM reduced repeated-output variability.} Legacy models showed
  17/19 positive SI effects (Hedges' \(g=1.18\), \(p=6.55\) \(\times\)
  10\(^{-5}\), parsed-field SI), NewModels showed 12/12 positive effects
  (\(g=3.08\)), and the combined random-effects estimate was
  \(g_{\text{RE}}\) = 1.09 {[}95\% CI 0.83, 1.35{]} across 31
  model-level pairs.
\item
  \textbf{State persistence reduced decision flipping in newer models.}
  In the NewModels cohort, DFR decreased by 82\% (0.385 \(\rightarrow\)
  0.069, \(g=1.52\)).
\item
  \textbf{The active mechanism was not JSON formatting alone.} The 4-arm
  ablation showed that value-only recomputation preserved the F/H/E
  effect (\(g=2.24\)) while eliminating the apparent JSON-only effect.
\item
  \textbf{Intrinsic model randomness under fixed anchor states was
  negligible.} FCS was approximately 1.0, localizing multi-turn
  instability mainly to state construction.
\item
  \textbf{CKM's advantage increased under sampling stochasticity.} At
  \(T=0.7\), the CKM effect reached \(g=2.87\).
\item
  \textbf{The sham-restriction ablation separated structure from
  epistemic content.} Structural scaffolding accounted for approximately
  45\% of the SI improvement, while reasoning-semantic F/H/E content
  accounted for approximately 55\%. CKM was the only arm that both
  improved SI and reduced DFR.
\end{enumerate}

These results do \textbf{not} show that CKM improves reasoning
correctness or decision quality. They show a narrower but important
result: behavioral consistency is measurable, differs substantially
across models, and can be partially improved by forcing models to
separate observed facts, inferred assumptions, and evaluative signals
before deciding. Code, prompts, and canonical datasets are available at
https://github.com/TeenyToolSoftware/cogos-behavioral-consistency.

\hypertarget{significance-statement}{%
\subsection{Significance Statement}\label{significance-statement}}

Most LLM benchmarks ask what a model can do. This paper asks a different
deployment-facing question: when the same situation returns, does the
model respond in a stable way, or does it drift and reverse its
decision?

The paper makes one bounded claim. In the tested Korean-language
decision scenarios, if an LLM is required to separate a decision input
into \textbf{Fact}, \textbf{Heuristic}, and \textbf{Emotion} before
choosing an action, then one form of behavioral instability becomes
measurable and can be partially reduced in repeated and state-persistent
trials.

The central contribution is therefore not a new LLM and not a claim of
better reasoning quality. It is a reproducible method for profiling and
partially stabilizing LLM behavior by making the model's reasoning state
explicit enough to measure repeated-response variability, decision
flips, and state drift.

\begin{center}\rule{0.5\linewidth}{0.5pt}\end{center}

\hypertarget{introduction}{%
\subsection{1. Introduction}\label{introduction}}

Large language models (LLMs) are usually evaluated by asking what they
can do: whether they solve benchmark questions, write correct code,
answer medical items, or pass standardized tests. This paper asks a
different question: \textbf{when the same situation is presented again,
does the model behave consistently?} A model can be highly capable and
still be unstable. It may produce semantically different answers to
identical prompts, preserve the same reasoning while changing its
action, or reverse a previous decision when its own prior output is
injected back as context.

This distinction matters for deployment. A medical assistant that gives
different triage advice for the same symptoms across sessions undermines
clinical trust. A legal assistant that changes its interpretation of the
same statute creates liability. A financial or engineering assistant
that changes its risk recommendation because the reasoning state was
represented differently introduces noise into decisions that require
stability. In these cases, the failure is not necessarily a lack of
capability. It is a failure of behavioral consistency.

We define \textbf{behavioral consistency} as the stability of a model's
semantic response, retained context, selected action, and propagated
state under repeated or state-persistent trials. This definition
deliberately separates consistency from correctness. A model can be
consistently wrong, and a model can be accurate on average while
unstable across repetitions. The present study measures the former
property and leaves decision quality as an independent question.

Our central hypothesis is simple: some LLM instability arises because
the model receives all information as undifferentiated text. In a human
decision situation, several kinds of information are mixed together:
observed facts, inferred assumptions, and affective or value-laden
signals. If these are not separated, a model may treat speculation as
fact, emotional emphasis as evidence, or a prior inference as an
observed constraint. We therefore test whether forcing the model to
separate information by epistemic role---\textbf{Fact},
\textbf{Heuristic}, and \textbf{Emotion} (F/H/E)---can reduce behavioral
variability.

To test this hypothesis, we develop the \textbf{Cognitive Kernel Model
(CKM)}. CKM is not a new base model and does not require fine-tuning. It
is a prompt-level state-enforcement framework that requires an LLM to
(i) classify input information into F/H/E slots, (ii) produce a decision
from that structured state, and (iii) optionally carry the structured
state forward into later turns. In plain terms, CKM asks the model to
keep a clean ledger of what is known, what is inferred, and what is
valued before it acts.

This framing differs from conventional prompt engineering.
Chain-of-Thought, Tree-of-Thought, and self-refinement methods structure
how a model reasons within a response {[}1--3, 14, 15{]}. CKM instead
structures what epistemic role each piece of information is allowed to
play and tests whether that enforced state structure changes repeated
behavior. Agent frameworks and memory systems can retain information
across turns {[}5--7, 38--43{]}, but retention alone does not specify
whether retained content is a fact, an inference, or an evaluative
signal. CKM addresses that categorization layer.

The study makes three main contributions, all within the claim boundary
that we measure consistency rather than correctness:

\begin{enumerate}
\def\labelenumi{\arabic{enumi}.}
\item
  \textbf{A behavioral consistency measurement suite.} We define six
  metrics---SI, CRR, DFR, FCS, ACS, and SDR---that capture different
  forms of response drift, decision flipping, and state instability.
  Four of these metrics (SI, DFR, FCS, SDR) are discriminative in the
  present experiments; CRR and ACS are reported as exploratory.
\item
  \textbf{A model-agnostic F/H/E state-enforcement framework.} CKM
  decomposes reasoning inputs into Fact, Heuristic, and Emotion roles
  and enforces that structure through a prompt-level schema. The
  framework can be applied to any text-input LLM without weight
  modification.
\item
  \textbf{Cross-vendor empirical profiling of LLM behavioral
  consistency.} Across 26 unique models from four vendors and two model
  generations, we show that baseline consistency varies substantially,
  CKM reduces output variability in most models, and state-persistence
  behavior differs sharply across architectures. Ablation studies show
  that the main active mechanism is epistemic role separation rather
  than JSON formatting alone.
\end{enumerate}

\textbf{Claim-chain boundary.} The paper uses one claim chain
throughout. First, LLM behavior varies across repeated presentations of
the same decision problem. Second, that variability can be measured with
repeated-trial and state-persistence metrics. Third, F/H/E state
enforcement reduces part of that variability in the tested setting.
Fourth, ablations indicate that the reduction is not explained by JSON
formatting alone. We do not claim that CKM improves factual accuracy,
ethical quality, or general reasoning capability.

Recent studies have begun to quantify LLM output variability directly:
Wang and Wang {[}34{]} found substantial inconsistency across 3.4
million finance-domain outputs under identical prompts, Atil et
al.~{[}61{]} demonstrated nondeterministic variability even under fixed
decoding parameters, and Haase et al.~{[}62{]} showed that within-model
variance components differ across task types. This paper builds on that
recognition but adds a structural intervention: instead of only
observing variability, we test whether a specific state organization can
reduce it.

The rest of the paper is organized as follows. Section 2 reviews related
work and identifies the gap addressed by CKM. Section 3 defines the CKM
architecture and F/H/E state transition. Section 4 defines the
behavioral consistency metrics. Section 5 describes the experimental
design. Section 6 reports the empirical results and ablations. Section 7
discusses mechanisms, limitations, and implications. Section 8
concludes.

\begin{center}\rule{0.5\linewidth}{0.5pt}\end{center}

\hypertarget{background-and-related-work}{%
\subsection{2. Background and Related
Work}\label{background-and-related-work}}

This section surveys three streams of prior work that together motivate
CKM and then identifies the specific gap we address. §2.1 reviews LLM
consistency evaluation and its single-dimensional limitations. §2.2
surveys cognitive architectures and structured reasoning frameworks from
psychology and cognitive science. §2.3 examines structured output,
memory, and state persistence approaches. §2.4 articulates the research
gap that CKM addresses, and §2.5 situates our contribution against
concurrent and adjacent work.

\hypertarget{llm-evaluation-capability-vs.-consistency}{%
\subsubsection{2.1 LLM Evaluation: Capability
vs.~Consistency}\label{llm-evaluation-capability-vs.-consistency}}

The dominant paradigm for LLM evaluation focuses on task performance,
measuring accuracy on standardized benchmarks (MMLU, HumanEval, GSM8K)
while implicitly treating each evaluation instance as independent
{[}26{]}. This design reflects a specific assumption: that measuring
what a model can do on diverse test items is sufficient for
characterizing its utility. Measuring how consistently it performs on
identical items requires a fundamentally different experimental design,
one based on repeated measurements rather than diverse sampling, and the
capability paradigm provides no infrastructure for this. Recent work on
LLM calibration {[}29{]}, prompt sensitivity {[}30{]}, and positional
bias {[}31{]} has begun to address aspects of this gap.

Shyr et al.~{[}33{]} propose repeatability metrics for medical LLM
evaluation, and Wang \& Wang {[}34{]} assess output consistency across
3.4 million finance outputs. These studies measure variability as an
observational phenomenon without structural intervention and employ
single-dimensional metrics that cannot detect the multi-dimensional
failure modes our data reveal (e.g., simultaneous high SI and high DFR).
Our six-metric suite addresses both limitations.

\hypertarget{cognitive-architectures-and-structured-reasoning}{%
\subsubsection{2.2 Cognitive Architectures and Structured
Reasoning}\label{cognitive-architectures-and-structured-reasoning}}

While these consistency measures capture variability as a phenomenon,
they do not explain its structural origins. Classical cognitive science
offers frameworks for structured reasoning that speak to this question.
Dual-process theory {[}8{]} posits fast and slow processing systems;
metacognition research {[}9{]} emphasizes self-monitoring; scaffolding
theory {[}10{]} demonstrates that external structure aids reasoning.
Computational models such as SOAR {[}16{]}, ACT-R {[}17{]}, and GOMS
{[}18{]} provide structured frameworks with procedural memory but lack
compatibility with modern LLM embeddings. Critically, these frameworks
provide structural proposals but lack quantitative evidence that their
structures improve behavioral consistency. Three examples illustrate
this gap.

SOAR separates modules by processing type without tagging epistemic
provenance. ACT-R organizes knowledge by representation format (chunks
vs.~production rules) rather than by information source. CoALA {[}44{]}
defines perception-action-memory loops without enforcing that perceived
facts, inferred hypotheses, and evaluative judgments be structurally
distinguished at the input stage. More recently, Kim {[}35{]} introduces
symbolic governance for neural reasoning and Kargupta et al.~{[}36{]}
map cognitive elements onto reasoning traces, advancing cognitive
structuring but without measuring whether such structuring yields
measurable consistency gains (Supplementary §S-RW for extended
discussion).

\hypertarget{structured-output-memory-and-state-persistence}{%
\subsubsection{2.3 Structured Output, Memory, and State
Persistence}\label{structured-output-memory-and-state-persistence}}

These cognitive frameworks propose structures but do not implement them
within LLM systems. Recent advances in structured LLM outputs, including
JSON-mode generation {[}12{]} and function calling {[}13{]}, provide
syntactic constraints that reduce formatting variability. CKM extends
this approach by enforcing not merely syntactic structure but semantic
role separation (Fact vs.~Heuristic vs.~Emotion), requiring no weight
modification {[}22{]}. The distinction matters because our ablation
data, confirmed by value-only artifact control, show that semantic
separation is the primary contributor to consistency improvement (§6.1).
Chain-of-Thought {[}1{]}, Tree-of-Thought {[}2{]}, Self-Refine {[}14{]},
and Graph-of-Thought {[}11{]} impose structure within isolated outputs
without cross-step consistency, leaving multi-turn instability
unaddressed.

A growing body of work addresses state persistence. A-MEM {[}38{]}
organizes memories via linking; Memoria {[}39{]} combines summarization
with knowledge graphs; Mem0 {[}40{]} provides production-grade
continuity; and MemoryOS {[}43{]} proposes an OS abstraction for agent
memory. These systems address what to retain over time, optimizing
retention measured via F1 or recall. By contrast, CKM addresses a
complementary question: how to categorize what is retained.

Memory systems and CKM operate on orthogonal dimensions; a system can
have perfect memory yet still conflate facts with inferences if it lacks
epistemic role separation, a prediction our ablation data confirm, where
format enforcement's measured SI contribution was attributable to
structural homogeneity rather than genuine semantic convergence, while
semantic separation produced robust consistency gains (§6.1). CKM does
not replace memory systems; it provides the epistemic categorization
layer that they lack. In a layered architecture, memory systems would
handle retention and retrieval while CKM would enforce that retrieved
content is classified by provenance before entering the reasoning
process.

None of the memory systems quantifies the behavioral impact of persisted
state on reasoning consistency. Our DFR and SDR metrics directly address
this gap, revealing that state persistence response varies dramatically
across models (DFR: 0.01--0.85) in ways invisible to retention-focused
evaluation.

\hypertarget{research-gap}{%
\subsubsection{2.4 Research Gap}\label{research-gap}}

The absence of a comprehensive consistency framework reflects two
structural factors. First, the dominance of the capability evaluation
paradigm has directed community resources toward diverse-item
benchmarking rather than repeated-measurement designs, leaving
consistency measurement tools underdeveloped. Second, the NLP and AI
communities have not adopted the epistemological distinction between
information types (observed, inferred, evaluated) as an organizing
principle, despite its long history in both Western epistemology
{[}47{]} and Indian pramana theory {[}52{]}. The categorization of
information by epistemic provenance has simply not been part of the
computational vocabulary.

As a result, the literature reviewed above leaves unaddressed the
simultaneous provision of (1) a multi-dimensional consistency
measurement suite, (2) explicit state representation with epistemic role
separation, (3) model-agnostic structural enforcement, (4) cross-vendor
behavioral profiling, and (5) quantification of state persistence impact
on downstream behavior. Table 1 summarizes this comparison.

\textbf{Table 1. Comparison with existing approaches.}

\begin{longtable}[]{@{}
  >{\raggedright\arraybackslash}p{(\columnwidth - 10\tabcolsep) * \real{0.2083}}
  >{\raggedright\arraybackslash}p{(\columnwidth - 10\tabcolsep) * \real{0.1083}}
  >{\raggedright\arraybackslash}p{(\columnwidth - 10\tabcolsep) * \real{0.1333}}
  >{\raggedright\arraybackslash}p{(\columnwidth - 10\tabcolsep) * \real{0.1833}}
  >{\raggedright\arraybackslash}p{(\columnwidth - 10\tabcolsep) * \real{0.1667}}
  >{\raggedright\arraybackslash}p{(\columnwidth - 10\tabcolsep) * \real{0.2000}}@{}}
\toprule\noalign{}
\begin{minipage}[b]{\linewidth}\raggedright
Feature
\end{minipage} & \begin{minipage}[b]{\linewidth}\raggedright
CoT/ToT {[}1,2{]}
\end{minipage} & \begin{minipage}[b]{\linewidth}\raggedright
Agent Fw. {[}5--7{]}
\end{minipage} & \begin{minipage}[b]{\linewidth}\raggedright
Struct. Output {[}12,13{]}
\end{minipage} & \begin{minipage}[b]{\linewidth}\raggedright
Memory Sys. {[}38--43{]}
\end{minipage} & \begin{minipage}[b]{\linewidth}\raggedright
\textbf{This Work}
\end{minipage} \\
\midrule\noalign{}
\endhead
\bottomrule\noalign{}
\endlastfoot
Consistency measurement & No & No & No & No & \textbf{6-metric suite} \\
State persistence & None & Implicit & None & Persistent &
\textbf{Explicit state injection} \\
Cross-model profiling & No & No & No & No & \textbf{26 models, 2
gen.} \\
F/H/E semantic separation & No & No & No & No & \textbf{Yes} \\
State impact measurement & No & No & No & No & \textbf{DFR/SDR} \\
\end{longtable}

\hypertarget{concurrent-and-adjacent-work}{%
\subsubsection{2.5 Concurrent and Adjacent
Work}\label{concurrent-and-adjacent-work}}

Several recent studies have independently investigated LLM output
consistency as a primary research concern, confirming the practical
importance of the phenomenon we address. Atil et al.~{[}61{]}
demonstrated that hosted LLM environments introduce nondeterministic
variability even under fixed parameters, proposing TARr@N and TARa@N
metrics for reproducibility assessment. Wang and Wang {[}34{]} conducted
3.4 million output comparisons across 50 repetitions in financial and
accounting domains, finding substantial inconsistency in factual tasks.
Shyr et al.~{[}33{]} adapted statistical repeatability and
reproducibility frameworks from metrology to evaluate medical diagnostic
LLM outputs.

Haase et al.~{[}62{]} decomposed within-LLM variance using intraclass
correlation coefficients, revealing that variance components differ
substantially across task types. Du et al.~{[}63{]} measured decision
flip rates under identical-prompt repetition in value-reasoning
contexts, finding that models reverse positions at nontrivial rates even
without external pressure.

While these studies individually establish that LLM consistency is a
measurable and consequential property, our work integrates four elements
that none addresses in combination: (a) epistemic provenance
decomposition as a designable state variable for consistency, (b) a
six-metric behavioral measurement suite capturing distinct failure
modes, (c) prompt-level state enforcement as an experimental
manipulation rather than merely an evaluation target, and (d)
cross-model behavioral profiling at the scale of 26 models from four
vendors with 37,403 observations. A recent survey {[}64{]} explicitly
notes the absence of a comprehensive consistency benchmark, a gap our
metric suite directly addresses.

We also distinguish our zero-perturbation reproducibility paradigm from
perturbation-based consistency evaluation, which measures response
stability under input variation (SCORE {[}65{]}; Zhang and Zhu
{[}66{]}); our design holds inputs constant and measures intrinsic
output variability.

A parallel line of work explores epistemic and cognitive classification
within LLM systems. NabaOS {[}67{]} applies Nyaya pramana categories
(pratyaksa, anumana, sabda, abhava) to implement claim-level provenance
classification with cryptographic verification in LLM agents, sharing
our grounding in Indian epistemological traditions. However, NabaOS
employs a four-category pramana taxonomy oriented toward hallucination
detection in tool-execution pipelines, whereas our F/H/E framework
integrates Western epistemology (Audi {[}47{]}) with Indian traditions
in a three-axis provenance classification designed for prompt-level
state enforcement and behavioral consistency profiling.

MOLES {[}68{]} classifies 11 epistemic stances that models adopt toward
their own knowledge claims; our framework classifies epistemic source
types of the information entering reasoning, operating at a different
level of analysis. Kargupta et al.~{[}36{]} propose a 28-element
cognitive taxonomy validated across 170K traces but do not implement
prompt-level enforcement or measure behavioral consistency under
structural intervention.

We note two terminological clarifications. First, the term ``Cognitive
Kernel'' has been independently used for an LLM agent autopilot system
(Zhang et al.~{[}69{]}); our Cognitive Kernel Model (CKM) is a
prompt-level epistemic state enforcement architecture with no
architectural overlap. Second, Pei et al.~{[}70{]} use the phrase
``behavioral fingerprinting of large language models''; their work
profiles behavioral styles and capabilities from single-trial
evaluations, whereas our behavioral profiling characterizes consistency
and reproducibility under repeated trials with cognitive state
manipulation.

Several recent preprints address adjacent questions, mostly in
\textbf{persona consistency / role-playing agent stability} rather than
input-side epistemic decomposition. Tosato et al.~{[}76{]} (PERSIST)
measure persona-conditioned behavior stability across model scale,
reasoning mode, and conversation history. Luo and Laban {[}78{]} (SPASM)
propose stable persona-driven agent simulation for multi-turn dialogue.
Liu et al.~{[}79{]} develop decoding-time persona importance estimation
for role-playing agents. Kim et al.~{[}80{]} (PICon) introduce
multi-turn interrogation for evaluating persona agent consistency. These
works probe whether assigned persona traits remain coherent across turns
and contexts; our setting carries no assigned persona and measures
behavioral consistency under repeated identical inputs, so the
contributions are complementary rather than overlapping.

A second adjacent strand examines \textbf{endpoint or latent-state
stability without persona conditioning}. Huang et al.~{[}75{]} document
failure modes of latent state persistence in frontier LLMs. Leshin et
al.~{[}77{]} build a behavioral-fingerprint Stability Monitor for
detecting LLM endpoint behavioral drift via output-distribution
comparison. Both characterize state or endpoint stability as a
monitoring target rather than as a property modifiable by a structured
input intervention.

A third strand evaluates \textbf{consistency under controlled input
variation}: Cavalin et al.~{[}81{]} (CAT) introduce metrics quantifying
the consistency-accuracy relation under controlled input variations.
Hsing {[}82{]} (MIRROR, ICLR 2026 withdrawn) proposes a modular
architecture for personalized multi-turn safety, sharing the
multi-turn-stability concern at the architectural-augmentation level.

None of these works treats an input-side epistemic decomposition as a
controlled experimental manipulation, nor do they jointly measure
repeated-trial consistency (SI) and state-persistent decision stability
(DFR) on a cross-vendor corpus of the present scale. Our framing differs
from PERSIST {[}76{]} and Pei et al.~{[}70{]} in a methodologically
substantive way: rather than profiling behavioral consistency as an
intrinsic, fixed property of each model, we treat it as a property
\textbf{modifiable by a prompt-level epistemic intervention}. F/H/E
decomposition thus enters the present design as a controlled independent
variable, not a profiling target.

\begin{center}\rule{0.5\linewidth}{0.5pt}\end{center}

\hypertarget{architecture-and-intervention-cognitive-kernel-model}{%
\subsection{3. Architecture and Intervention: Cognitive Kernel
Model}\label{architecture-and-intervention-cognitive-kernel-model}}

This section defines the intervention tested in the experiments. The
\textbf{Cognitive Kernel Model (CKM)} is not introduced as a new base
model, cognitive architecture, or software kernel. In this paper, CKM is
a model-agnostic prompt-level state-enforcement method for LLM decision
tasks (Figure 1). It does three things:

\begin{enumerate}
\def\labelenumi{\arabic{enumi}.}
\tightlist
\item
  it separates input information into \textbf{Fact}, \textbf{Heuristic},
  and \textbf{Emotion} roles;
\item
  it requires the model to produce an output that respects those roles;
  and
\item
  when needed, it carries the resulting structured state into later
  turns.
\end{enumerate}

CKM is therefore both an intervention and a measurement aid. As an
intervention, it constrains how the model organizes information before
deciding. As a measurement aid, it exposes whether the model preserves
that organization across repeated and multi-turn trials. CKM does not
modify model weights. It treats each LLM as a black-box semantic
processor and imposes structure externally through prompt-level
enforcement. Throughout the paper, ``CKM'' refers to this tested F/H/E
state-enforcement method unless a broader, explicitly marked extension
is being discussed.

The motivation is straightforward. Standard prompts ask the model to
interpret a mixed text input directly. CKM instead asks the model to
first answer three bookkeeping questions: What is actually given? What
is being inferred? What evaluative or priority signal is present? The
hypothesis tested in this paper is that this bookkeeping reduces
ambiguity in the reasoning state and thereby reduces behavioral
variability.

This prompt-level design is a methodological choice rather than a
convenience. The central question is whether the same external state
structure can improve consistency across many vendors and architectures.
Fine-tuning would mix the state-enforcement effect with model-specific
training effects and would make cross-vendor comparison difficult.
Prompt-level enforcement functions more like a runtime discipline: it
constrains the interface through which reasoning is expressed, while
leaving the underlying model unchanged. We use this analogy only to
locate the intervention; CKM is implemented as a structured prompt
schema, not as operating-system software.

The remainder of this section develops CKM in four steps. §3.1
formalizes the kernel state and transition function. §3.2 describes the
Tri-Engine extension, which is included for architectural completeness
but not evaluated in this paper. §3.3 specifies the prompt-level schema
used in the experiments. §3.4 explains why explicit state persistence is
required for multi-turn consistency.

\hypertarget{kernel-state-and-transition-function}{%
\subsubsection{3.1 Kernel State and Transition
Function}\label{kernel-state-and-transition-function}}

The kernel state is the structured record that CKM asks the model to
maintain. The term ``kernel'' is used in the limited sense of a minimal
state record, not as a claim that the prompt acts like an
operating-system kernel. It is also not intended to be a full model of
human cognition. It is a minimal operational state for decision
consistency: a record of what the model treats as given, what it treats
as inferred, and what it treats as evaluative or priority-relevant.

At time \(t\), CKM represents this state as:

\[S_t = \{F_t, H_t, E_t\}\]

where \(F_t\) is the \textbf{Fact} vector (observable or verifiable
information), \(H_t\) is the \textbf{Heuristic} vector (inferred
relations, assumptions, or uncertain interpretations), and \(E_t\) is
the \textbf{Emotion} vector (affective tone, salience, or value/priority
signals).\footnote{\textbf{Competing-interests disclosure.} Components
  of the CKM framework described in this paper are the subject of patent
  applications filed by an author of this paper. These span both the
  mechanisms evaluated here---the F/H/E epistemic decomposition and the
  separation of epistemic roles---and broader specification elements that
  are not experimentally validated in this study, namely additional
  unvalidated state slots beyond the core F/H/E decomposition; multi-axis
  persistence and hierarchical restoration of the cognitive state across
  sessions; and a fourth \emph{testimony} (\(T\)) provenance axis
  extending the F/H/E scheme (§7.6). The applications are pending and not
  yet published, and identifiers are therefore omitted. This disclosure
  is made in the interest of transparency; the experimental design,
  results, and their interpretation are reported independently of this
  intellectual-property interest.}

\hypertarget{why-separate-at-all}{%
\paragraph{Why separate at all?}\label{why-separate-at-all}}

The F/H/E decomposition addresses a specific failure mode: when
reasoning systems process factual observations, speculative inferences,
and affective evaluations as undifferentiated input, they conflate
verified information with tentative hypotheses and allow emotional
emphasis to contaminate logical structure. Classical epistemology
identifies this confusion, namely treating observed facts and
speculative inferences as epistemically equivalent, as a primary source
of reasoning instability (Audi {[}47{]}), and the problem is amplified
in LLMs because all inputs arrive as homogeneous text tokens. Epistemic
provenance, the information about where a claim originates and how it
was derived, is maintained in biological cognition through distinct
sensory, inferential, and interoceptive channels.

LLM token sequences carry no such labels, erasing these distinctions
entirely. Without external enforcement of these distinctions, models
generally lack explicit mechanisms to prevent fact-inference conflation,
and the resulting ambiguity is associated with increased output
variability. CKM imposes separation at the input stage precisely to
reduce this ambiguity.

\hypertarget{why-fact-heuristic-and-emotion-specifically}{%
\paragraph{Why Fact, Heuristic, and Emotion
specifically?}\label{why-fact-heuristic-and-emotion-specifically}}

Given that separation is beneficial, the question becomes which
categories to separate. The three axes are organized by epistemic
provenance, classifying information by how it arrived rather than by
processing speed {[}8{]}, functional domain {[}58{]}, or phenomenal
modality. This criterion, while standard in epistemology {[}47, 48{]}
and codified in Indian pramana theory as pratyaksa (perception) and
anumana (inference) {[}52{]}, has not been adopted as an organizing
principle in any computational cognitive architecture. Each axis serves
an irreducible function. \(F\) grounds reasoning in verifiable
observations, helping to prevent hallucination. \(H\) marks the boundary
between known and assumed, helping to prevent unwarranted certainty.

\(E\) encodes priority and value signals that guide action selection.
This role for emotion is consistent with Damasio's somatic marker
hypothesis {[}49{]}, Tappolet's account of emotions as evaluative
perceptions {[}50{]}, and de Sousa's argument that emotions solve the
frame problem by directing attention to what matters {[}51{]}. We draw
on the shared insight from these traditions that emotion carries
decision-relevant information, while acknowledging that CKM
operationalizes this insight through explicit tagging rather than the
implicit somatic influence these authors describe.

A two-axis system (\(F\)+\(H\) only) would leave affective signals
untagged, allowing emotional emphasis to infiltrate factual and
inferential processing without explicit marking, a pattern we term
epistemic contamination. Adding a fourth or fifth axis is theoretically
possible. For example, Indian Nyaya epistemology recognizes testimony
(sabda) as an independent source {[}52{]}, and Mimamsa traditions
enumerate up to six channels. In this study, we treat three as a
functionally sufficient partition for the present validation scope, and
we note the testimony dimension as a natural extension in §7.6.

\hypertarget{what-is-novel-about-this-decomposition}{%
\paragraph{What is novel about this
decomposition?}\label{what-is-novel-about-this-decomposition}}

While the individual components of this partition have precedents in
philosophy and cognitive science, their specific combination is novel.
The specific novelty lies in three simultaneous moves that no reviewed
framework performs together. First, CKM splits the traditional
``cognition'' category into perception-as-given (\(F\)) and
inference-as-derived (\(H\)), a separation that Hilgard's
cognition-affection-conation trilogy {[}58{]} leaves implicit. Second,
it elevates emotion from a modulator or output, as treated in
dual-process theory {[}8{]} and predictive processing {[}59{]}, to an
input-level information channel with mandatory tagging.

Third, it organizes all three by epistemic provenance rather than by
processing speed (Kahneman's System 1/System 2 {[}8{]}), control level
(Stanovich's reflective/algorithmic/autonomous {[}60{]}), or
motivational conflict (Plato's tripartite soul). The closest precedent
is the Social Information Processing model {[}53, 54{]}, which
integrates cognitive, emotional, and behavioral components, but that
framework is specific to social situations; CKM operates as a
domain-general epistemic state layer. Importantly, we do not claim that
F/H/E is the unique decomposition, nor that it models human cognition;
we claim it is a functionally sufficient partition for reducing output
variability, and our data support this claim.

\hypertarget{why-is-this-separation-especially-important-for-llms}{%
\paragraph{Why is this separation especially important for
LLMs?}\label{why-is-this-separation-especially-important-for-llms}}

The importance of explicit separation becomes clear when we consider
what LLMs lack. For biological cognizers, epistemic provenance is
partially maintained by distinct neural pathways: sensory channels tag
perceptual input, metacognitive processes flag inferential uncertainty,
and interoceptive signals carry affective information {[}49, 55{]}. LLMs
lack these architectural separations. All inputs, whether sourced from
verified databases, speculative reasoning, or value-laden descriptions,
arrive as undifferentiated token sequences. This homogenization means
that without external structural enforcement, the model may process a
verified statistic and an unsubstantiated inference with identical
computational status.

In practical terms, token sequences carry no explicit labels indicating
whether a statement is verified or merely inferred. CKM provides the
external enforcement that the architecture itself cannot supply,
functioning as an epistemic provenance layer above the model.

\hypertarget{empirical-confirmation}{%
\paragraph{Empirical confirmation}\label{empirical-confirmation}}

The preceding arguments establish the theoretical rationale for F/H/E
separation; we now show that these design choices also receive empirical
support from our data. The ablation study (§6.1) confirms that F/H/E
semantic separation is the primary contributor to consistency
improvement, with value-only analysis confirming that format
enforcement's measured SI effect reflects structural homogeneity rather
than semantic convergence (§6.1, Supplementary §S-Robustness). The
2.4-fold larger CKM effect in ambiguity-dominant scenarios observed in
the Legacy cohort (§6.3) is consistent with the prediction that F/H/E
separation most directly addresses fact-inference conflation, which is
precisely the condition that ambiguity scenarios impose.

In the primary NewModels cohort (n = 12) this differentiation collapses
to a 1.07-fold ratio (ambiguity \(\Delta\)SI +0.0416 vs.~non-ambiguity
+0.0443; §6.3), consistent with §7.2's discussion of cohort-specific
mechanism weighting. The 98.4\% LLM valid rate across 26 models (§3.3)
suggests that the decomposition is practically realizable without
model-specific adaptation, and the human pilot (§7.3) found perfect
F/H/E compliance in the manually reviewed CKM subset (0 violations / 15
annotated responses; §S11.2), providing preliminary indication that the
separation is compatible with existing cognitive categorization
abilities, though demand characteristics in a small sample cannot be
ruled out.

The preceding evidence addresses whether F/H/E separation works; a
separate question is whether the choice of three axes is principled or
arbitrary. An analogy from color science clarifies the point but
requires refinement: RGB is not merely an engineering convenience but
reflects the biological structure of human trichromatic vision.
Similarly, F/H/E is not arbitrary but grounded in epistemologically
established cognitive role separations. However, just as RGB is not the
only color space (HSV and CMYK serve different purposes), F/H/E is not
the only possible decomposition; it is one optimized for reasoning
stability, and we make no stronger claim.

With the F/H/E partition established, we now define how the kernel state
evolves over time. CKM-Core evolves according to a deterministic state
transition function:

\[S_{t+1} = \delta(S_t, \Omega_{x,t}, \Omega^{\text{Engine}}_t)\]

where \(\Omega_{x,t}\) is the normalized input bundle and
\(\Omega^{\text{Engine}}_t\) represents aggregated engine outputs.
Determinism here refers only to the formal state-update rule: given
identical initial state and input sequence, the resulting represented
state is uniquely determined. It is not a claim that the underlying LLM
computation is deterministic across hosted API calls.

\hypertarget{tri-engine-architecture}{%
\subsubsection{3.2 Tri-Engine
Architecture}\label{tri-engine-architecture}}

CKM-Core handles state and decomposition; a separate architectural layer
can process this normalized state through multiple specialized engines.
The CKM framework can be extended with a multi-engine processing
architecture (\(\Psi\)-Architect, DecisionEngine, Meta-Structure) that operates
on the normalized Kernel State; this extension is not evaluated in the
present study and is left as a direction for future work (§7.6).

\includegraphics{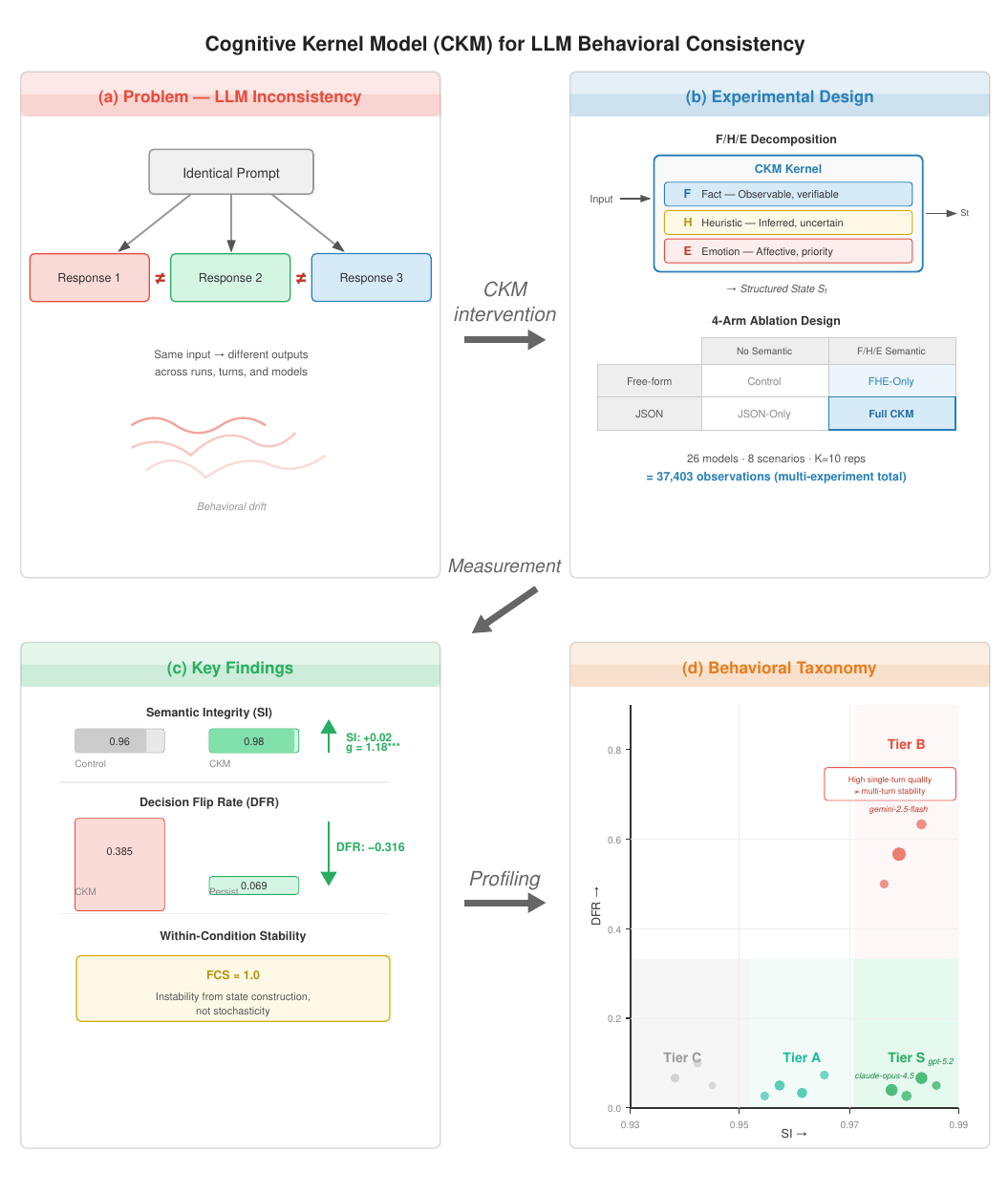}

\textbf{Figure 1. Cognitive Kernel Model (CKM) for LLM Behavioral
Consistency.} Four-panel conceptual overview: (a) Problem: LLM
consistency variability on identical inputs; (b) Experimental Design:
F/H/E decomposition, 4-arm ablation, 26 LLM models; (c) Key Findings: under CKM, SI
increases relative to free-form control (with the largest gains in models
with the lowest baseline consistency, a behavioral-variance reducer
pattern; see §6.3), and DFR decreases sharply under state persistence
(single-turn CKM vs.\ persisted CKM); (d) Behavioral Taxonomy:
Tier S/A/B/C model archetypes distributed in the DFR-SI behavioral space
(see §6.3 for tier definitions).

\hypertarget{ckm-as-prompt-level-enforcement}{%
\subsubsection{3.3 CKM as Prompt-Level
Enforcement}\label{ckm-as-prompt-level-enforcement}}

In implementation, CKM is simply a structured instruction given to the
model before it answers. The model is not asked merely to format its
answer as JSON; it is asked to classify the information it uses before
choosing an action. The standard implementation used in this study is a
5-slot JSON schema:

\begin{lstlisting}
{
  "fact": ["observable, verifiable information"],
  "heuristic": ["inferred relations (marked with '(inferred)')"],
  "emotion": {"code": "E3", "intensity": 4},
  "action": {"code": "A1"},
  "reason": "decision rationale linking F/H/E to chosen action"
}
\end{lstlisting}

This schema was applied uniformly across all 26 models. LLMs achieved a
98.4\% valid rate (24,054/24,443 LLM-only observations across two
generations), and the full-corpus valid rate including the Exp1 human
pilot is 96.7\% (24,092/24,923); Exp1's 38 valid of 480 designed
responses reflects human participant workload rather than data quality
issues (§7.3, Supplementary §S11). These rates confirm that LLMs can
reliably comply with F/H/E requirements without model-specific
adaptation. Enforcement operates through two mechanisms: schema
compliance forcing explicit slot filling, and semantic role separation
requiring verifiable facts together with uncertainty-marked heuristics
and coded emotions.

CKM is realized through prompt-level enforcement; alternative
architectural configurations exist but are beyond the scope of this
study. All quantitative results in §6 are based on the standard F/H/E
implementation.

\hypertarget{why-state-persistence-is-necessary}{%
\subsubsection{3.4 Why state persistence is
necessary}\label{why-state-persistence-is-necessary}}

LLMs are fundamentally stateless: each API call resets the computation,
and no mechanism inherently carries reasoning context across turns.
Commercial chat interfaces create an appearance of continuity by
re-injecting prior conversation history into each new context window or
by maintaining separate memory modules, but the underlying model itself
retains no state between calls; it processes each context window as if
encountering the conversation for the first time. Recent empirical work
confirms that frontier LLMs fail to maintain latent state across turns
even when explicitly instructed to do so {[}75{]}, providing independent
justification for external prompt-level enforcement.

Decision consistency in multi-turn interactions therefore requires
explicit state injection, because without it, each turn begins from a
blank slate and the model has no basis for maintaining coherence with
prior outputs. The CKM\_PERSIST condition tests this by injecting Turn 1
output as structured context for Turn 2. The results reveal that state
persistence response is itself a diagnostic tool: some models achieve
DFR near zero under injection while others exhibit DFR near 0.85,
exposing a model-intrinsic property that is largely invisible without
state persistence measurement (§6.2).

Progressive stabilization under chain protocols (§6.2) provides
preliminary evidence that iterated state enforcement can drive the
system toward more stable behavioral attractors.

\begin{center}\rule{0.5\linewidth}{0.5pt}\end{center}

\hypertarget{behavioral-consistency-metrics}{%
\subsection{4. Behavioral Consistency
Metrics}\label{behavioral-consistency-metrics}}

This section defines the measurement side of the claim chain. CKM is
evaluated with metrics designed for repeated behavior, not single-answer
accuracy. Standard NLP metrics such as BLEU, ROUGE, and F1 require
reference answers and therefore do not fit open-ended decision scenarios
where no single correct answer exists. Conversely, a simple
repeatability score is not enough: a model can keep similar wording
while changing its action, or preserve an action while drifting
semantically.

We therefore measure behavioral consistency along several axes. The main
question behind each metric is deliberately simple:

\begin{itemize}
\tightlist
\item
  \textbf{SI:} Do repeated answers to the same input stay semantically
  close?
\item
  \textbf{CRR:} Does the model retain the relevant facts from the
  scenario?
\item
  \textbf{DFR:} Does the model change its selected action when prior
  state is injected?
\item
  \textbf{FCS:} Under fixed anchor states, does the model repeatedly
  choose the same action?
\item
  \textbf{ACS:} How concentrated or dispersed are the selected actions
  across repetitions?
\item
  \textbf{SDR:} Does the structured F/H/E state survive propagation
  across turns?
\end{itemize}

The independence of these axes is part of the empirical claim. In §6.3,
for example, some models show high single-turn SI but high multi-turn
DFR, meaning that single-turn semantic stability does not guarantee
decision stability under state persistence. This is why the paper
reports a metric suite rather than a single consistency score.

\hypertarget{semantic-integrity-si}{%
\subsubsection{4.1 Semantic Integrity
(SI)}\label{semantic-integrity-si}}

We begin with the most basic consistency question: given identical
inputs, do the model's responses converge semantically? SI quantifies
this convergence.

SI measures semantic similarity of responses to identical prompts across
\(K\) repetitions:

\[SI(m, s, c) = \frac{1}{K} \sum_{i=1}^{K} \cos(\mathbf{e}_i, \bar{\mathbf{e}})\]

where \(\mathbf{e}_i = \text{SentenceBERT}(\text{response}_i)\) and
\(\bar{\mathbf{e}} = \frac{1}{K}\sum_{j=1}^{K} \mathbf{e}_j\) is the
L2-normalized centroid of the \(K\) response embeddings. We use
\passthrough{\lstinline!paraphrase-multilingual-mpnet-base-v2!}
(768-dim), selected through sensitivity analysis showing that
English-optimized alternatives produce ceiling compression on Korean
text, yielding null effect sizes (\(d=0.05\); see Supplementary §S1). To
control for the possibility that JSON schema elements (recurring keys
such as \passthrough{\lstinline!"fact"!},
\passthrough{\lstinline!"heuristic"!}) inflate cosine similarity
through structural homogeneity rather than semantic convergence, we
computed a value-only SI variant (SI\('\)) by stripping all JSON keys
and structural syntax before embedding.

The value-only SI preserved the CKM effect (\(g=2.88\), 100.8\%
retention; Supplementary §S-Robustness), confirming that the measured
consistency improvement reflects semantic content rather than format
artifacts.

\textbf{Interpretation boundary.} SI is an embedding-based operational
proxy for within-framework semantic convergence, not a direct measure of
reasoning correctness or decision quality. High SI indicates that a
model's responses cluster in embedding space under a given prompt
structure (a necessary but not sufficient condition for accurate
reasoning). Convergence toward shared error remains possible, and SI
should be interpreted alongside behavioral metrics (DFR, SDR) and the
claim boundaries stated in §1.

\hypertarget{context-retention-rate-crr}{%
\subsubsection{4.2 Context Retention Rate
(CRR)}\label{context-retention-rate-crr}}

While SI captures same-turn convergence, CRR addresses whether prior
context is preserved across turns, a prerequisite for any multi-turn
stability claim.

\[CRR = \frac{|\{f_j \in F_{\text{gold}} \mid \exists\, s_i \in R,\ \text{sim}(s_i, f_j) > \theta\}|}{|F_{\text{gold}}|}\]

where \(F_{\text{gold}}\) is the set of mandatory scenario facts and
\(\theta = 0.75\). CRR captures information loss through the reasoning
process.

\hypertarget{decision-flip-rate-dfr}{%
\subsubsection{4.3 Decision Flip Rate
(DFR)}\label{decision-flip-rate-dfr}}

Semantic similarity can mask decision instability: a model may produce
similar text while reversing its conclusion. DFR isolates the
decision-level changes that SI cannot detect.

\[DFR(m) = \frac{|\{(i,j) : a_{T1,i} \neq a_{T2,j}\}|}{N_{\text{matched}}}\]

Lower DFR indicates greater decision stability under state persistence.
Our DFR measures binary decision changes across repeated identical
trials without external pressure, distinguishing it from adversarial
flip rates that measure capitulation under challenge {[}71, 72{]},
value-reasoning flip rates in constrained domains {[}63{]}, and
confirmation-bias reversal rates under counter-evidence exposure
{[}73{]}.

\hypertarget{few-shot-coherence-score-fcs}{%
\subsubsection{4.4 Few-shot Coherence Score
(FCS)}\label{few-shot-coherence-score-fcs}}

Beyond single-trial stability, FCS measures how consistent a model's
responses are across multiple repetitions of the same prompt, capturing
coherence under repeated sampling.

\[FCS(m, s) = \frac{\max_a |\{k : a_k = a\}|}{K}\]

Under fixed anchor states, FCS quantifies intrinsic model randomness.

\hypertarget{action-consistency-score-acs}{%
\subsubsection{4.5 Action Consistency Score
(ACS)}\label{action-consistency-score-acs}}

Beyond semantic and decision consistency, ACS captures whether the
action path selected across repetitions remains stable. This matters
when downstream systems consume only the recommended action rather than
the full response.

\[ACS = 1 - \frac{H(\mathbf{p})}{\log_2 |A|}\]

where \(H(\mathbf{p})\) is the Shannon entropy of the action
distribution. This behavioral trial-to-trial measure is distinct from
cross-modal text-visual fidelity scores that share the ACS abbreviation
in recent GUI agent literature {[}74{]}. In the present scenario set,
CRR and ACS showed limited discriminative power: CRR was compressed
against its upper bound because the key-fact inventory for each scenario
is largely recoverable from the prompt itself, while ACS varied in a
narrow range around 0.65--0.70 with no consistent CKM/Control ordering
(Exp2 scenario-level means: ACS = 0.69 Control vs.~0.65 CKM). Neither
metric discriminated models or conditions in ways that altered the
primary findings.

The quantitative analyses in this paper therefore rely on SI, DFR, FCS,
and SDR, which showed clear discriminative power across the model
cohort (Table 2). Per-scenario CRR and ACS values are archived with the canonical
datasets for completeness (Supplementary §S15).

\hypertarget{semantic-drift-resistance-sdr}{%
\subsubsection{4.6 Semantic Drift Resistance
(SDR)}\label{semantic-drift-resistance-sdr}}

State persistence is only effective if the kernel state itself is
preserved across turns. SDR measures how much of the F/H/E content
survives a state injection.

\[SDR(m) = \frac{1}{N_{\text{matched}}} \sum_{(i,j)} \cos(\mathbf{e}_{T1,i}^{FH}, \mathbf{e}_{T2,j}^{FH})\]

Higher SDR indicates greater state preservation across turns. Our SDR
quantifies intrinsic semantic drift across repeated identical trials,
distinct from adversarial context-switching evaluations that employ
superficially similar terminology under perturbation paradigms.

\textbf{Table 2. Metric summary.}

\begin{longtable}[]{@{}
  >{\raggedright\arraybackslash}p{(\columnwidth - 6\tabcolsep) * \real{0.0845}}
  >{\raggedright\arraybackslash}p{(\columnwidth - 6\tabcolsep) * \real{0.3099}}
  >{\raggedright\arraybackslash}p{(\columnwidth - 6\tabcolsep) * \real{0.2676}}
  >{\raggedright\arraybackslash}p{(\columnwidth - 6\tabcolsep) * \real{0.3380}}@{}}
\toprule\noalign{}
\begin{minipage}[b]{\linewidth}\raggedright
Metric
\end{minipage} & \begin{minipage}[b]{\linewidth}\raggedright
Measures
\end{minipage} & \begin{minipage}[b]{\linewidth}\raggedright
Scale
\end{minipage} & \begin{minipage}[b]{\linewidth}\raggedright
Failure Mode
\end{minipage} \\
\midrule\noalign{}
\endhead
\bottomrule\noalign{}
\endlastfoot
\textbf{SI} & Semantic variability & 0--1 (\(\uparrow\)) & Response
drift \\
\textbf{CRR} & Information retention & 0--1 (\(\uparrow\)) & Context
loss \\
\textbf{DFR} & Decision instability & 0--1 (\(\downarrow\)) & State
sensitivity \\
\textbf{FCS} & Action consistency & 0--1 (\(\uparrow\)) & Action
randomness \\
\textbf{ACS} & Population convergence & 0--1 (\(\uparrow\)) &
Inter-agent disagreement \\
\textbf{SDR} & State preservation & 0--1 (\(\uparrow\)) & Cognitive
drift \\
\end{longtable}

\begin{center}\rule{0.5\linewidth}{0.5pt}\end{center}

\hypertarget{experimental-design}{%
\subsection{5. Experimental Design}\label{experimental-design}}

This section explains how the consistency claims were tested. The design
has three constraints. First, the intervention must be evaluated across
vendors, because CKM is claimed to be model-agnostic. Second, each
scenario must be repeated enough times to measure variability rather
than isolated performance. Third, the mechanism must be separated from
superficial formatting, because an apparent consistency gain could
otherwise be caused by JSON structure alone.

Accordingly, §5.1 summarizes the experiment set and rationale. §5.2
documents the models and scenarios. §5.3 defines the four core
conditions: Control, CKM, CKM\_PERSIST, and Chain. §5.4 summarizes the
statistical methods; full derivations are provided in Supplementary
§S12.

\hypertarget{overview}{%
\subsubsection{5.1 Overview}\label{overview}}

We conducted four core experiments, a 4-arm ablation study, a 5-arm
sham-restriction ablation, and a temperature robustness probe. Core
experiments were replicated on a NewModels cohort (13 models collected,
2025--2026; 12 primary after exclusion of gpt-4.1-mini, see below),
yielding 37,403 observations across two generations (35,475 in the
primary 12-model cohort plus 1,928 from gpt-4.1-mini reported in the
§6.6 case study) (Table 3). The experimental design reflects several deliberate
choices. The 4-vendor, 2-generation structure is necessary because the
model-agnostic claim requires evidence that effects are not
vendor-specific artifacts; a single-vendor study would leave this
confound unresolved.

The \(K=10\) repetition count balances centroid stability for SI
computation (centroids computed from fewer than 5 repetitions are
unreliable) against API cost constraints across 26 models and 8
scenarios. The 4-arm ablation (§6.1) was designed specifically to
preempt the critique that CKM effects reduce to ``advanced prompt
engineering'' by orthogonally decomposing format enforcement and
semantic role separation in a 2×2 factorial design.

\textbf{Table 3. Experimental overview.}

\begin{longtable}[]{@{}
  >{\raggedright\arraybackslash}p{(\columnwidth - 8\tabcolsep) * \real{0.1839}}
  >{\raggedright\arraybackslash}p{(\columnwidth - 8\tabcolsep) * \real{0.2184}}
  >{\raggedright\arraybackslash}p{(\columnwidth - 8\tabcolsep) * \real{0.2184}}
  >{\raggedright\arraybackslash}p{(\columnwidth - 8\tabcolsep) * \real{0.1839}}
  >{\raggedright\arraybackslash}p{(\columnwidth - 8\tabcolsep) * \real{0.1954}}@{}}
\toprule\noalign{}
\begin{minipage}[b]{\linewidth}\raggedright
Exp
\end{minipage} & \begin{minipage}[b]{\linewidth}\raggedright
Target
\end{minipage} & \begin{minipage}[b]{\linewidth}\raggedright
Comparison
\end{minipage} & \begin{minipage}[b]{\linewidth}\raggedright
\(N_{\text{obs}}\)
\end{minipage} & \begin{minipage}[b]{\linewidth}\raggedright
Primary Metrics
\end{minipage} \\
\midrule\noalign{}
\endhead
\bottomrule\noalign{}
\endlastfoot
\textbf{Exp1} & Human (\(n=6\)) & Control vs CKM & 480 & SI (proxy),
ACS \\
\textbf{Exp2} & LLM (19 Legacy) & Control vs CKM & 3,040 & SI, CRR,
FCS \\
\textbf{Exp2-NM} & LLM (12 NM primary) & Control vs CKM & 1,920 & SI \\
\textbf{Exp3} & LLM (19 Legacy) & CKM vs CKM\_PERSIST &
4,670\(^\dagger\) & SI, DFR, SDR \\
\textbf{Exp3-NM} & LLM (12 NM primary) & CKM vs CKM\_PERSIST & 3,840 &
SI, DFR, SDR \\
\textbf{Exp3 Extra} & LLM (16 Legacy) & T1→T2→T3 Chain & 2,655 & SI,
FCS, DFR, SDR \\
\textbf{Exp3X-NM} & LLM (12 NM primary) & T1→T2→T3 Chain & 2,014 & SI,
DFR, SDR \\
\textbf{Study A} & LLM (6 Legacy) & 4-Arm Ablation & 768+728 & SI \\
\textbf{Study B} & LLM (12 NM primary) & 4-Arm Ablation & 3,840 & SI \\
\textbf{Exp6} & LLM (12 NM primary) & 5-Arm Sham Ablation &
11,520\(^\S\) & SI, DFR, SDR \\
\textbf{Primary Total} & & & \textbf{35,475} & \\
\textbf{Full corpus\(^\P\)} & & & \textbf{37,403} & \\
\end{longtable}

\(^\S\) Exp6 collected 3 new sham arms (Sham-A/B/C, 3,840 each for 12
primary models = 11,520 new records). Vanilla and CKM data reused from
Exp2-NM and Exp3-NM respectively.

\(^\P\) Full corpus (37,403) = 35,475 primary + 1,928 gpt-4.1-mini
observations collected with the 13-model set but excluded from primary
analysis due to vendor deprecation (see §5.2, §6.6). The case study data
are reported in §6.6 and the sensitivity analysis (§S-Sensitivity).
Primary total (35,475) reflects the 12-model active cohort
(\passthrough{\lstinline!ablation\_newmodels\_metrics\_v1.2\_active12.json!},
\passthrough{\lstinline!exp3extra\_newmodels\_metrics\_v1.2\_active12.json!})
used in all primary analyses.

\(^\dagger\) Exp3 \(N=4\),670 is the design-level matched-pair count.
The canonical analysis dataset (SHA256:
\passthrough{\lstinline!16d04208...!}) contains Exp3 records within a
combined Exp3/Exp3Extra file (\(N=7\),479); metrics are computed on the
Exp3-tagged subset. grok-2-latest (33/240 records, S1--S2 only) was
excluded from analysis due to incomplete collection.

\(^\ddagger\) Legacy cohort composition: 21 total unique models
collected across vendors. Exp2 and Exp3 each analyze 19 Legacy models;
Exp3 Extra analyzes 16 Legacy models; Study A (ablation) analyzes 6
Legacy models. The ``17/19 positive effects'' in the Abstract refers to
Exp2's \(n=19\) subset: 17 models showed positive \(\Delta\)SI; the two
exceptions (claude-3-haiku, claude-haiku-4-5) showed near-zero or
slightly negative \(\Delta\)SI. The 26-model headline combines 21 Legacy
+ 13 NewModels (collected) − 8 overlapping models in cross-generation
comparison. The primary NewModels analysis uses 12 models (gpt-4.1-mini
excluded); the combined analysis pools 19 Legacy + 12 NewModels = 31
model-level pairs.

\hypertarget{models-and-scenarios}{%
\subsubsection{5.2 Models and Scenarios}\label{models-and-scenarios}}

Our cross-generation, cross-vendor claim requires a principled model
selection; we describe the resulting 26-model cohort and the 8-scenario
protocol derived from pilot testing.

We selected 26 unique models spanning four vendors across two
generations: a Legacy cohort (21 models, 2024--2025) and a NewModels
cohort (13 models collected, 2025--2026, spanning 4 vendors \(\times\) 3
tiers: Flagship/Standard/Fast). We define two NewModels sub-cohorts:

\begin{itemize}
\tightlist
\item
  \textbf{NewModels-12 (primary)}: Excludes gpt-4.1-mini. OpenAI issued
  a deprecation notice for gpt-4.1-mini on 2026-02-13, overlapping with
  our data collection window (2026-02-15 to 2026-03-11). Independent
  reports documented instruction-following degradation during this
  period. All primary aggregate statistics, per-model tables, and
  inferential tests in §6 use this 12-model cohort.
\item
  \textbf{NewModels-13 (sensitivity)}: Includes gpt-4.1-mini. Reported
  in the case study (§6.6) and sensitivity analysis (Supplementary
  §S-Sensitivity) to demonstrate that all directional findings are
  preserved regardless of inclusion.
\end{itemize}

The exclusion combines two vendor-side criteria documented before data
inspection, namely (1) vendor-issued deprecation notice during the
collection window and (2) independent reports of instruction-following
degradation, with a third post-hoc anomaly-detection criterion (3)
identified after data collection: an anomalous behavioral profile
relative to the cohort, namely the sole model in which DFR\_persist
\textgreater{} DFR\_ckm. Two further criteria govern the reporting
protocol rather than the exclusion decision itself: (4) removal does not
change the direction of any primary finding (verified via the
§S-Sensitivity 13-model re-analysis after the exclusion was applied),
and (5) full transparency via the §6.6 case study and sensitivity
reporting.

We do not characterize this five-element procedure as a pre-registered
analysis plan; criteria (3) and (4) are explicitly outcome-dependent and
were operationalized after the data were observed. We retain the
exclusion because criteria (1)--(2) constitute vendor-side evidence that
the gpt-4.1-mini observations may reflect a transient deprecation-window
state rather than a stable model, and we report both the 12-model
primary results and the 13-model sensitivity results so that readers can
apply their own interpretive weighting.

Eight overlapping models enable cross-generation comparison. Full model
lists are provided in Supplementary §S22 (NewModels) and §S-App (Legacy
coverage matrix).

Ten scenarios were initially designed, spanning ambiguity resolution,
ethical conflict, resource allocation, and error handling. Two scenarios
(S7 and S8) were discarded after pilot testing revealed insufficient
ambiguity to elicit meaningful response variability; the original
numbering was preserved for traceability with data collection logs. S6
was initially excluded due to apparent near-uniform responses in pilot
data, but re-examination of the full dataset revealed sufficient
discriminative variance across conditions, prompting its retention in
the main analysis. The remaining eight scenarios (S1--S6, S9, S10) were
used across all experiments (Supplementary §S4 for full texts).

The student daily-life domain was selected because it provides decision
situations where no objectively correct answer exists, forcing models to
exercise judgment rather than recall factual knowledge. Domains with
clear correct answers (e.g., medical diagnosis with known conditions,
mathematics) would confound consistency measurement with accuracy
measurement. Student scenarios naturally juxtapose factual constraints,
uncertain inferences, and affective considerations, making them
well-suited for testing whether F/H/E decomposition improves reasoning
stability under conditions where all three information types co-occur.

\hypertarget{conditions}{%
\subsubsection{5.3 Conditions}\label{conditions}}

With models and scenarios fixed, the remaining design dimension is the
experimental manipulation. Four conditions isolate structural
enforcement from state persistence, and both from multi-turn chaining.

Four experimental conditions were employed. In the Control condition,
models produced free-form responses with no structural enforcement. In
the CKM condition, models produced JSON-structured responses with
mandatory F/H/E fields. The CKM\_PERSIST condition followed a two-turn
protocol in which Turn 1 output was injected as context for Turn 2. The
Chain condition extended this to a three-turn protocol (T1 anchor, T2
branch \(\times\) 10, T3 chain \(\times\) 10).

\hypertarget{statistical-methods}{%
\subsubsection{5.4 Statistical Methods}\label{statistical-methods}}

Given the small-sample, cross-model structure of our design, standard
parametric tests alone are insufficient; we combine parametric and
nonparametric methods with bootstrap confidence intervals.

The primary unit of analysis was model-level means paired across
conditions. We used paired \(t\)-tests (two-tailed) and permutation
tests (10,000 resamples, seed = 42) for robustness, with Hedges' \(g\)
effect sizes accompanied by 2,000-resample bootstrap 95\% CIs (seed =
42). Legacy effect sizes are reported as Hedges' \(g\) throughout for
cross-cohort comparability (Cohen's \(d\) values, where also computed,
are provided in Supplementary §S-Exp2).

Multiple comparison correction followed the Holm-Bonferroni procedure
applied within each pre-defined confirmatory hypothesis family: (a) the
4-arm ablation contrasts (H1, H2, H3a, H3b, H3-synergy; family size 5;
Table 6); (b) the 5-arm sham-restriction pairwise comparisons computed
separately for SI and DFR\(_{\text{persist}}\) (10 pairwise tests per
metric; Table 11); and (c) the per-metric paired contrasts within each
NewModels experiment (e.g., DFR/SI/SDR for Exp3-NM, family size 3; Table
7). Cross-table corrections are not applied; tier-classification
analyses, scenario-level breakdowns, and the behavioral-variance reducer
analysis are reported as exploratory and not Holm-corrected. Robustness
checks included Wilcoxon signed-rank tests and logit-transformed
\(t\)-tests (Supplementary §S12 for full statistical methods).

\begin{center}\rule{0.5\linewidth}{0.5pt}\end{center}

\hypertarget{results}{%
\subsection{6. Results}\label{results}}

The results follow the same claim chain introduced in §1. §6.1 asks
whether CKM changes repeated-output consistency and then decomposes the
effect into format and semantic components. §6.2 asks whether explicit
state persistence changes decision flipping and chain stability. §6.3
treats the resulting SI--DFR patterns as behavioral profiles rather than
capability scores. §6.4 reports robustness checks. §6.5 reports the
sham-restriction ablation, which separates structural scaffolding from
epistemic content. §6.6 presents an excluded-model case study for
transparency.

The results are organized around four questions.

\begin{enumerate}
\def\labelenumi{\arabic{enumi}.}
\tightlist
\item
  \textbf{Does F/H/E state enforcement reduce repeated-output
  variability?} (§6.1)
\item
  \textbf{Is the effect caused by epistemic role separation, or merely
  by structured formatting?} (§6.1 and §6.5)
\item
  \textbf{Does carrying structured state forward reduce decision
  flipping across turns?} (§6.2)
\item
  \textbf{Do different model families have distinct behavioral
  consistency profiles?} (§6.3--§6.6)
\end{enumerate}

All results should be read within the paper's claim boundary: the
experiments measure consistency, not correctness. A higher SI or lower
DFR indicates greater stability under the tested protocol, not
necessarily a better decision.

\hypertarget{ckm-effectiveness-and-mechanism}{%
\subsubsection{6.1 CKM Effectiveness and
Mechanism}\label{ckm-effectiveness-and-mechanism}}

We begin with the primary effectiveness question (does CKM improve
consistency?) and then decompose the mechanism via the 4-arm ablation.
Structural compliance results come first; the ablation isolates which
component of CKM does the work.

\hypertarget{structural-compliance-exp2}{%
\paragraph{Structural Compliance
(Exp2)}\label{structural-compliance-exp2}}

Exp2 tested whether CKM's structural enforcement produces measurable SI
gains under standard conditions across a broad Legacy cohort.

Across 19 Legacy models (\(N=3\),040), CKM enforcement increased SI by
+0.016 (Hedges' \(g=1.18\), \(p=6.55\) \(\times\) 10\(^{-5}\),
parsed-field SI), with 17/19 models showing positive \(\Delta\)SI (Figure 2). SI
was computed from parsed-field text (concatenation of the
\passthrough{\lstinline!fact!} list and
\passthrough{\lstinline!reason!} string extracted from the JSON
response), which is inherently free of JSON structural artifact by
construction (see §4.1 and Supplementary §S-Robustness for the NewModels
raw-output cohort, where an explicit artifact control was applied).

The raw-output basis (Option B in §S-Exp2) yields \(g=0.61\),
\(p=0.012\), 14/19 positive; both bases are reported in §S-Exp2 as a
sensitivity comparison, and parsed-field is adopted as primary
throughout this paper to maintain a single SI substrate consistent with
§4.1.

\textbf{Table 4. Aggregate SI results by cohort.}

\begin{longtable}[]{@{}
  >{\raggedright\arraybackslash}p{(\columnwidth - 8\tabcolsep) * \real{0.3298}}
  >{\raggedright\arraybackslash}p{(\columnwidth - 8\tabcolsep) * \real{0.2021}}
  >{\raggedright\arraybackslash}p{(\columnwidth - 8\tabcolsep) * \real{0.1064}}
  >{\raggedright\arraybackslash}p{(\columnwidth - 8\tabcolsep) * \real{0.1170}}
  >{\raggedright\arraybackslash}p{(\columnwidth - 8\tabcolsep) * \real{0.2447}}@{}}
\toprule\noalign{}
\begin{minipage}[b]{\linewidth}\raggedright
Cohort
\end{minipage} & \begin{minipage}[b]{\linewidth}\raggedright
\(N_{\text{models}}\)
\end{minipage} & \begin{minipage}[b]{\linewidth}\raggedright
\(\Delta\)SI
\end{minipage} & \begin{minipage}[b]{\linewidth}\raggedright
Effect size
\end{minipage} & \begin{minipage}[b]{\linewidth}\raggedright
\(p\)-value
\end{minipage} \\
\midrule\noalign{}
\endhead
\bottomrule\noalign{}
\endlastfoot
Legacy (Exp2) & 19 & +0.016 & \(g=1.18\) & 6.55 \(\times\)
10\(^{-5}\) \\
NewModels-12 (Exp2-NM, primary) & 12 & +0.021 & \(g=3.08\) &
\(1.82 \times 10^{-7}\) \\
\textbf{Combined (Legacy + NM-12)} & \textbf{31} & \textbf{+0.018} &
\textbf{\(g=1.53\)} & \textbf{\(9.0 \times 10^{-10}\)} \\
\end{longtable}

Random-effects meta-analysis across all 31 model-level effects yields
\(g_{\text{RE}}\) = 1.09, 95\% CI {[}0.83, 1.35{]}, \(I^2\) = 77.6\%,
indicating a robust effect with substantial between-model heterogeneity.
The Combined-row \(\Delta\)SI = +0.018 in Table 4 is computed as the
unweighted mean of the 31 model-level \(\Delta\)SI values (each model
contributing one effect), not as the cohort-size-weighted average (which
would be (19 \(\times\) 0.016 + 12 \(\times\) 0.021) / 31 = 0.0179,
agreeing with the unweighted figure to three decimal places).

Practical-significance qualifier: SI is bounded above by 1.0, and the
Control baseline in this study clusters around SI \(\approx\) 0.95
(Legacy) and 0.95 (NM), leaving a measurement headroom of approximately
0.05 absolute units; the observed \(\Delta\)SI of +0.016 (Legacy
parsed-field) and +0.021 (NM-12) thus correspond to roughly 32\% and
42\% of the available headroom, respectively. Effect-size interpretation
in Hedges' \(g\) should be read against the small within-pair variance
characteristic of model-level paired designs, which inflates \(g\)
relative to per-trial designs at the same absolute \(\Delta\).

The primary NewModels replication (12/12 positive, \(g=3.08\))
substantially exceeded the Legacy effect (\(g=1.18\) parsed-field), and
among 8 overlapping models, 7/8 showed larger \(\Delta\)SI in the newer
cohort. Tier-level analysis revealed that Flagship models (\(\Delta\)SI
= +0.025, \(n=4\)) showed the largest gains, followed by Fast (+0.021,
\(n=4\)) and Standard (+0.017, \(n=4\)). Per-model results are in
Supplementary §S-Exp2.

\includegraphics{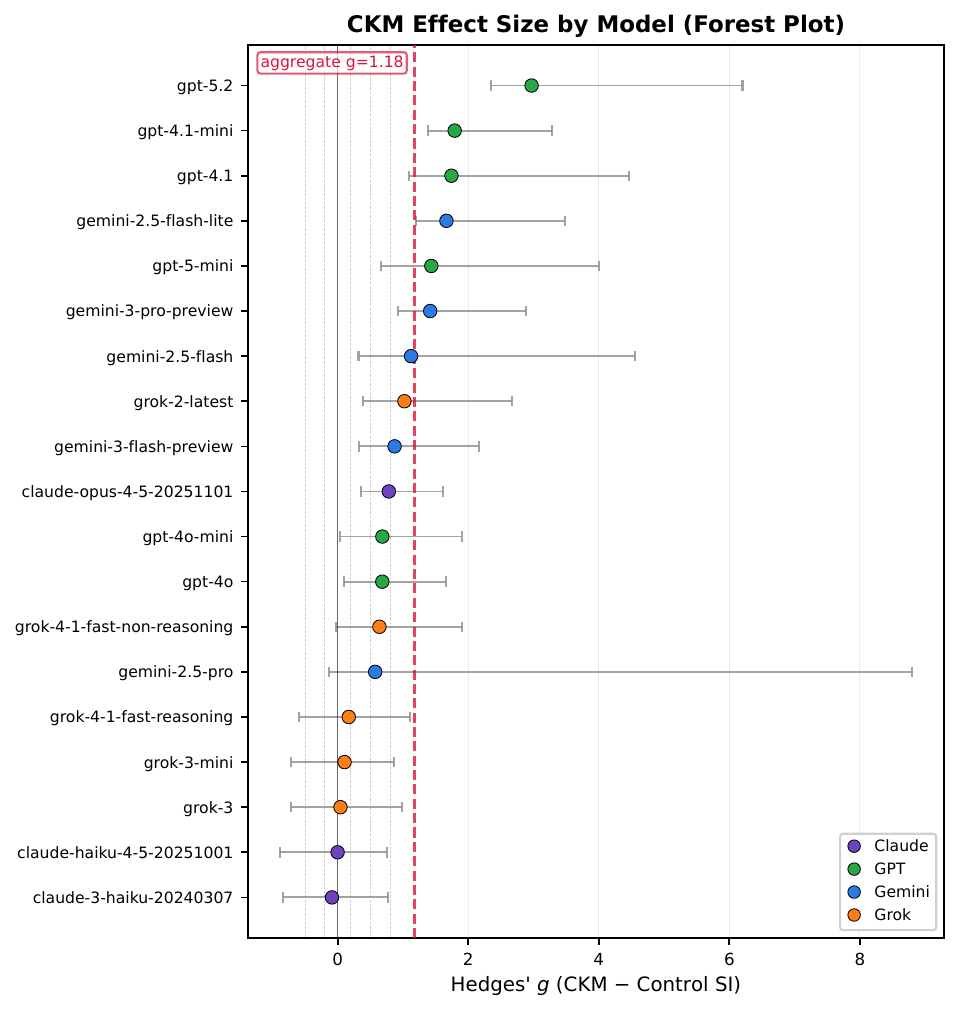}

\textbf{Figure 2. Hedges' \(g\) forest plot for 19 Legacy models.} CKM
effect size with 95\% bootstrap CI, ordered by effect magnitude. The
dashed vertical line marks the aggregate effect (\(g=1.18\),
parsed-field SI). GPT family models show the largest positive effects;
Claude family models show near-zero or slightly negative effects.

\hypertarget{ablation-format-vs.-semantic-separation-study-b}{%
\paragraph{Ablation: Format vs.~Semantic Separation (Study
B)}\label{ablation-format-vs.-semantic-separation-study-b}}

Effectiveness without mechanism decomposition would leave open the
critique that any measured gain reflects JSON format rather than
epistemic role separation. Study B orthogonally decomposes these two
factors.

The 4-arm ablation (\(N=3\),840, 12 primary NewModels; gpt-4.1-mini
excluded from primary cohort and analyzed separately in §6.6 case study)
crossed format enforcement (JSON vs.~free-form) with semantic role
separation (F/H/E vs.~none) (Table 5). Statistics below are computed from
\passthrough{\lstinline!ablation\_newmodels\_metrics\_v1.2\_active12.json!}
(n = 12).

\textbf{Table 5. Ablation arm design and results.}

\begin{longtable}[]{@{}llll@{}}
\toprule\noalign{}
Arm & Format & Semantic & Mean SI \\
\midrule\noalign{}
\endhead
\bottomrule\noalign{}
\endlastfoot
Control & Free-form & None & 0.960 \\
JSON-Only & JSON & None & 0.977 \\
F/H/E-Only & Free-form & F/H/E & 0.981 \\
Full CKM & JSON & F/H/E & 0.985 \\
\end{longtable}

\textbf{Table 6. Study B planned contrasts (Holm-corrected, n = 12
paired models).}

\begin{longtable}[]{@{}
  >{\raggedright\arraybackslash}p{(\columnwidth - 6\tabcolsep) * \real{0.3587}}
  >{\raggedright\arraybackslash}p{(\columnwidth - 6\tabcolsep) * \real{0.2500}}
  >{\raggedright\arraybackslash}p{(\columnwidth - 6\tabcolsep) * \real{0.1196}}
  >{\raggedright\arraybackslash}p{(\columnwidth - 6\tabcolsep) * \real{0.2717}}@{}}
\toprule\noalign{}
\begin{minipage}[b]{\linewidth}\raggedright
Hypothesis
\end{minipage} & \begin{minipage}[b]{\linewidth}\raggedright
Comparison
\end{minipage} & \begin{minipage}[b]{\linewidth}\raggedright
Hedges' \(g\)
\end{minipage} & \begin{minipage}[b]{\linewidth}\raggedright
\(p_{\text{Holm}}\)
\end{minipage} \\
\midrule\noalign{}
\endhead
\bottomrule\noalign{}
\endlastfoot
H1: Format effect & JSON vs Control & \textbf{1.875}\(^\dagger\) &
\(1.16 \times 10^{-4}\) *** \\
H2: Semantic effect & F/H/E vs Control & \textbf{2.637} &
\(5.32 \times 10^{-6}\) *** \\
H3a: Semantic over format & CKM vs JSON & \textbf{1.263} &
\(2.59 \times 10^{-3}\) ** \\
H3b: Format over semantic & CKM vs F/H/E & \textbf{1.231} &
\(2.59 \times 10^{-3}\) ** \\
Combined effect (artifact-driven) & CKM vs max(JSON, F/H/E) &
\textbf{0.835} & \(1.99 \times 10^{-2}\) * \\
\end{longtable}

\(^\dagger\) Value-only recomputation (Supplementary §S-Robustness)
revealed that H1's measured effect (\(g=1.875\)) reflects JSON
structural homogeneity rather than semantic convergence (value-only
\(g=-0.20\), ns). H2's effect was fully preserved (value-only
\(g=2.24\), \(p<10^{-6}\)). Original statistics are retained for
transparency; see §S-Robustness Table S-R3 for corrected effect sizes.

Under the original SI measurement, both format enforcement and F/H/E
semantic separation showed significant effects (Table 6). However,
value-only recomputation, which strips all JSON keys and structural
syntax before embedding, revealed a critical dissociation: the F/H/E
semantic separation effect was fully preserved (\(g=2.24\),
\(p<10^{-6}\)), while the JSON-only format effect was eliminated
(\(g=-0.20\), ns; Supplementary §S-Robustness). This indicates that H1's
measured contribution (\(g=1.87\)) reflected JSON structural homogeneity
inflating cosine similarity rather than genuine semantic convergence.

The ablation's primary contribution is therefore the isolation of F/H/E
semantic separation as the active mechanism: the FHE-Only condition
(free-form output with semantic role separation, no JSON enforcement)
achieved SI = 0.981, significantly exceeding Control (0.960) without any
format confound. Full CKM significantly exceeded FHE-Only (H3b:
\(g=1.231\), \(p_{\text{Holm}}\) = 2.59 \(\times\) 10\(^{-3}\)), but
this residual advantage is attributable to JSON structural homogeneity
rather than semantic synergy, as confirmed by the value-only analysis.
The pilot study (Study A, 6 models, \(N=768\)) showed consistent
directionality (Supplementary §S-Ablation).

\hypertarget{bridging-interpretation}{%
\paragraph{Bridging Interpretation}\label{bridging-interpretation}}

The value-only analysis clarifies the ablation's mechanistic
implications. The original 2×2 design was intended to orthogonally
decompose format and semantic contributions. Under standard SI
measurement, both appeared significant, suggesting independent variance
sources. However, the value-only recomputation reveals that format
enforcement's measured effect operated primarily at the measurement
level: JSON schema elements (recurring keys, structural syntax) inflated
Sentence-BERT cosine similarity through structural homogeneity rather
than semantic convergence. When this measurement artifact is removed,
the semantic separation effect stands alone as the primary mechanism.

This reinterpretation does not imply that format enforcement has no
behavioral effect on models. JSON slot-filling may still constrain model
behavior in ways not captured by SI, for example by ensuring complete
coverage of required reasoning components or by reducing response length
variability. However, format enforcement's contribution to measured
semantic convergence is not distinguishable from structural homogeneity
within the current measurement framework. The critical finding is that
F/H/E semantic separation produces genuine semantic convergence (SI\('\)
= 0.980 vs.~Control 0.960) even in the absence of any format enforcement
(FHE-Only condition), establishing epistemic role separation as the
sufficient condition for the observed consistency improvement.

\hypertarget{state-persistence-and-chain-stability}{%
\subsubsection{6.2 State Persistence and Chain
Stability}\label{state-persistence-and-chain-stability}}

Single-turn consistency is necessary but not sufficient; a consistency
framework must also handle multi-turn dynamics. We now examine state
persistence as a single-step injection effect and then extend to chain
protocols.

\hypertarget{state-persistence-exp3}{%
\paragraph{State Persistence (Exp3)}\label{state-persistence-exp3}}

Exp3 tests whether injecting Turn 1 output as Turn 2 context stabilizes
decisions or instead triggers re-evaluation, with per-model DFR
revealing how different architectures respond to injected state.

In Legacy (19 models), DFR ranged from 0.01 to 0.85, a dramatic
between-model difference that reveals state persistence response as a
model-intrinsic property (Figure 3). Three factors plausibly drive high DFR values.
First, reasoning-mode activation: some models (particularly those with
explicit chain-of-thought mechanisms) appear to treat injected state as
a new problem to be solved rather than as authoritative context,
triggering full re-evaluation from first principles. Second, context
window competition: when the injected prior state competes with the
current prompt for attention, models with shorter effective context
windows or weaker long-range attention may preferentially attend to the
immediate instruction.

Third, helpfulness optimization: RLHF training that rewards ``improved''
answers may incentivize models to revise rather than preserve prior
conclusions, interpreting state persistence as an opportunity for
correction rather than a constraint on consistency.

In the primary NewModels cohort (\(N=3\),840, 12 models), state
persistence reduced DFR by 82\%:

\textbf{Table 7. Exp3-NM aggregate results (primary, 12 models).}

\begin{longtable}[]{@{}lllll@{}}
\toprule\noalign{}
Metric & CKM & CKM\_PERSIST & Hedges' \(g\) & \(p\) \\
\midrule\noalign{}
\endhead
\bottomrule\noalign{}
\endlastfoot
DFR & 0.385 & 0.069 & \textbf{1.522} &
\textbf{\(1.4 \times 10^{-4}\)} \\
SI & 0.974 & 0.966 & −0.896 & 0.007 \\
SDR & 0.939 & 0.943 & 0.130 & 0.637 ns \\
\end{longtable}

All 12 primary NewModels showed DFR improvement under state persistence,
and 6 achieved perfect decision preservation (DFR = 0), compared to 0/19
in Legacy. This generational shift indicates that newer models are
better calibrated for state persistence. Fast models showed the lowest
mean DFR (0.011), followed by Flagship (0.019) and Standard (0.177; the
elevated Standard mean is driven by gemini-2.5-pro at 0.644, see §7.5
limitation 8 and per-model breakdown in Supplementary §S-Exp3).

The decrease in SI under state persistence does not indicate harm;
rather, it reflects models actively responding to injected state, a form
of contextual engagement. SDR of approximately 0.94 confirmed that
semantic content was preserved, while what changed was the decision
derived from that content. This pattern reveals a consistency-stability
tradeoff: state persistence improved decision stability (lower DFR) at
the cost of slightly reduced semantic homogeneity. Per-model breakdowns
are in Supplementary §S-Exp3.

\includegraphics{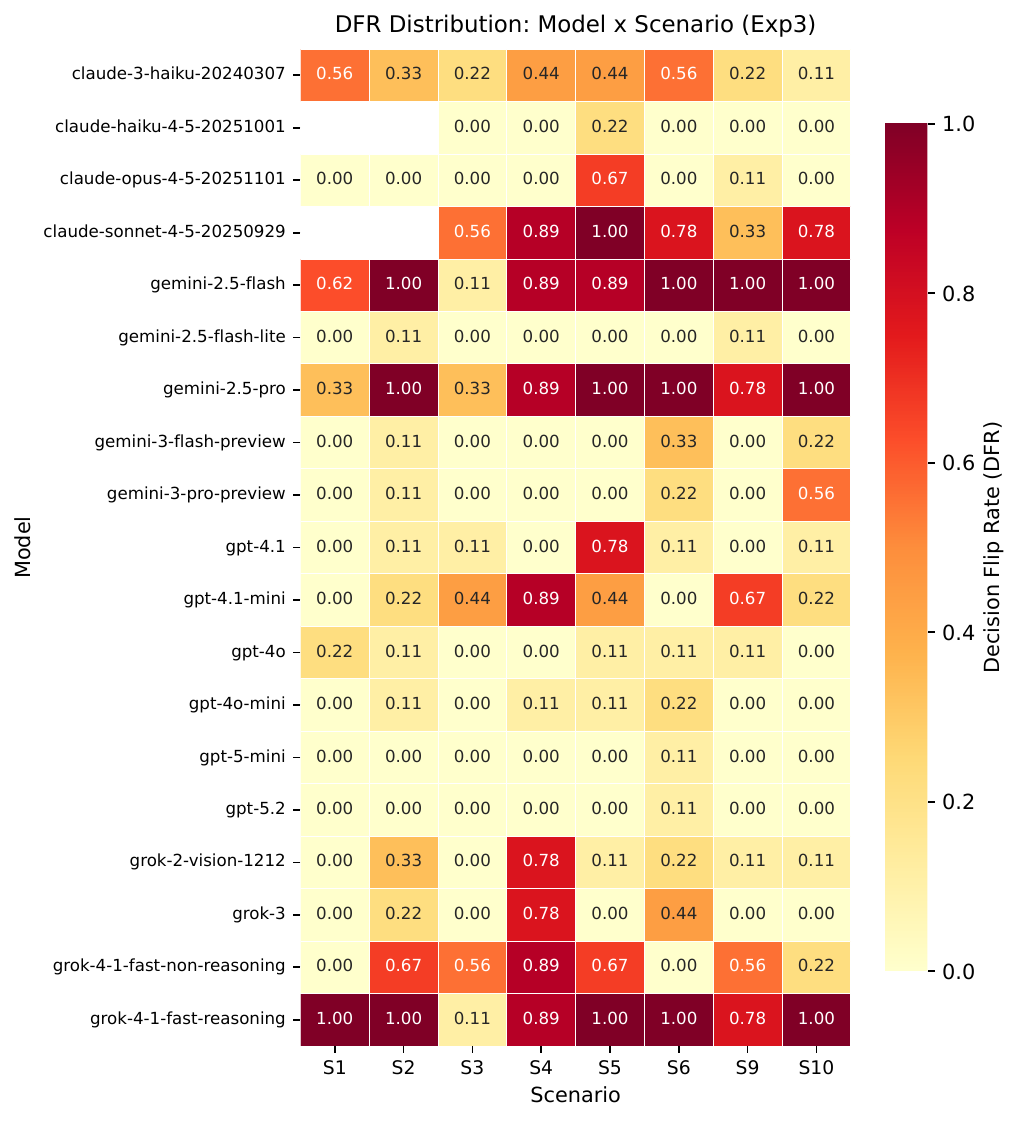}

\textbf{Figure 3. Decision Flip Rate (DFR) by model under state
persistence.} Stability group coloring: green (DFR \textless{} 0.15,
stable), yellow (0.15--0.50, moderate), red (\textgreater{} 0.50,
unstable). DFR spans from 0.01 to 0.85 across the 19 Legacy models shown.

\hypertarget{chain-stability-exp3-extra}{%
\paragraph{Chain Stability (Exp3
Extra)}\label{chain-stability-exp3-extra}}

A single injection tests whether state persistence works once; iterated
injection tests whether the effect accumulates or decays.

Under three-turn chain protocols, mean DFR decreased from 0.328 (T1→T2)
to 0.289 (T2→T3) while SDR increased from 0.992 to 0.995 across 16
Legacy models, demonstrating progressive stabilization consistent with
attractor dynamics. In the primary NewModels cohort, 6/12 achieved
perfect temporal stability (DFR = 0 across both transitions), compared
to 2/16 in Legacy.

FCS of approximately 1.0 under fixed anchor states demonstrated that
multi-turn instability originates from state construction and injection,
not from model stochasticity (Supplementary §S-Exp3X for full chain
results).

\hypertarget{cross-experiment-behavioral-profiles}{%
\subsubsection{6.3 Cross-Experiment Behavioral
Profiles}\label{cross-experiment-behavioral-profiles}}

The preceding sections establish that CKM improves single-turn
consistency and that state persistence stabilizes multi-turn decisions.
We now examine what patterns emerge when these dimensions are combined
across models.

\hypertarget{tier-classification}{%
\paragraph{Tier Classification}\label{tier-classification}}

Individual metrics isolate specific failure modes; combining them
reveals behavioral archetypes that single metrics cannot show.
Integrating SI (single-turn quality) and DFR (multi-turn stability)
yields a behavioral taxonomy:

\textbf{Table 8. Behavioral tier taxonomy.} Four-tier partition by
single-turn SI and multi-turn DFR\(_{\text{persist}}\). Tiers are
mutually exclusive and collectively exhaustive; cutoffs (SI = 0.95/0.97,
DFR = 0.15) reflect observed gaps in the joint distribution and are
treated as descriptive for the present cohort (§S-Profiles for
cutoff-sensitivity).

\begin{longtable}[]{@{}
  >{\raggedright\arraybackslash}p{(\columnwidth - 4\tabcolsep) * \real{0.0482}}
  >{\raggedright\arraybackslash}p{(\columnwidth - 4\tabcolsep) * \real{0.4337}}
  >{\raggedright\arraybackslash}p{(\columnwidth - 4\tabcolsep) * \real{0.5181}}@{}}
\toprule\noalign{}
\begin{minipage}[b]{\linewidth}\raggedright
Tier
\end{minipage} & \begin{minipage}[b]{\linewidth}\raggedright
Criteria
\end{minipage} & \begin{minipage}[b]{\linewidth}\raggedright
Representative Models
\end{minipage} \\
\midrule\noalign{}
\endhead
\bottomrule\noalign{}
\endlastfoot
\textbf{S} & SI \textgreater{} 0.97 and DFR \textless{} 0.15 &
claude-opus-4-6, gpt-5.4, grok-3 \\
\textbf{A} & 0.95 \textless{} SI \(\leq\) 0.97 and DFR \textless{} 0.15
& gpt-5-mini, gemini-3.1-pro-preview, gpt-5.2 \\
\textbf{B} & SI \textgreater{} 0.95 and DFR \(\geq\) 0.15 &
gemini-2.5-flash, gemini-2.5-pro \\
\textbf{C} & SI \(\leq\) 0.95 & gpt-4o, grok-2, claude-3-haiku \\
\end{longtable}

Tiers S and A partition the ``stable, low-flip'' region by single-turn
quality; Tier B captures the adequate-single-turn-quality but
elevated-flip pattern, spanning moderate (DFR 0.15--0.50) to severe (DFR
\textgreater{} 0.50) decision instability; Tier C groups
lower-single-turn-quality models regardless of flip behavior. The
criteria are both mutually exclusive and collectively exhaustive by
construction: every model with SI measurement is classified into exactly
one tier (Table 8).

The cutoff values were chosen post-hoc to land near observed gaps in the
joint SI--DFR distribution within our cohort: SI = 0.97 separates the
upper concentration of frontier models from the broader 0.95--0.97 band,
SI = 0.95 separates this band from the long lower tail, and DFR = 0.15
reflects an approximate elbow between models with near-zero flip rates
and those with persistent multi-turn instability.

We treat these thresholds as a descriptive partition for the present
cohort rather than as universal cutoffs; small perturbations (e.g., SI
cutoffs at 0.96 or 0.98) reassign one or two models near the boundary
without altering the qualitative four-tier structure (sensitivity in
Supplementary §S-Profiles). All Tier-classification analyses are
reported as exploratory and are not Holm-corrected (§5.4).

The Tier B classification reported above uses CKM\_PERSIST DFR from Exp3
(Legacy) and Exp3-NM (NewModels-12) --- the same 24-unique-model
cross-cohort plane shown in Figure 4 --- as the multi-turn stability
axis (gemini-2.5-flash's DFR = 0.82 originates from the Legacy Exp3
measurement); alternative stability axes (e.g., chain DFR from Exp3X)
yield the same qualitative tiering with \(\leq\) 1 model boundary
reassignment.

Tier B models present a distinctive pattern. gemini-2.5-flash achieved
high single-turn SI (Exp2 CKM SI = 0.972; per-scenario range as in
Supplementary Table S3) yet exhibited Exp3 DFR = 0.82 with CKM\_PERSIST
SI = 0.967, demonstrating that high single-turn quality does not
guarantee multi-turn stability. This phenomenon is invisible to standard
benchmarks, which do not distinguish between single-turn and multi-turn
consistency.

\includegraphics{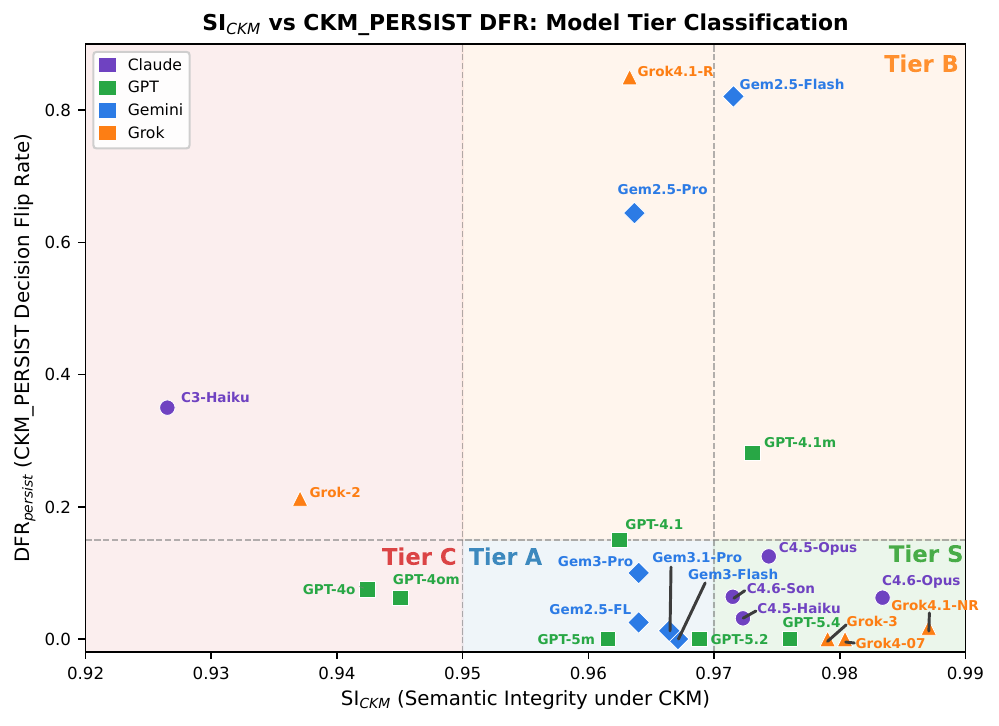}

\textbf{Figure 4. SI \(\times\) DFR behavioral quadrant.} The 24-model
union pool (19-model Legacy Exp2/Exp3 cohort combined with the 13-model
NewModels cohort, 8 overlapping models using NewModels data); the 23
models with both SI and CKM\_PERSIST DFR measurements are plotted
(grok-3-mini omitted for lacking a DFR measurement) and mapped into four
mutually exclusive and collectively
exhaustive tiers: S (SI \textgreater{} 0.97 and DFR \textless{} 0.15), A
(0.95 \textless{} SI \(\leq\) 0.97 and DFR \textless{} 0.15), B (SI
\textgreater{} 0.95 and DFR \(\geq\) 0.15), and C (SI \(\leq\) 0.95).
DFR axis = CKM\_PERSIST DFR (Exp3 / Exp3-NM). The SI (\(x\)) axis is truncated (begins at 0.92, not 0) for legibility. Tier B spans both
moderate-flip (DFR 0.15--0.50) and severe-flip (DFR \textgreater{} 0.50)
regimes; the upper-right subregion contains models with high single-turn
consistency but high multi-turn decision instability.

\hypertarget{behavioral-variance-reducer-effect}{%
\paragraph{Behavioral-Variance Reducer
Effect}\label{behavioral-variance-reducer-effect}}

Tier classification partitions models discretely; a continuous
relationship with baseline consistency provides complementary
information. Beyond tier classification, CKM exhibited a systematic
relationship with baseline performance: in the Legacy cohort, models
with lower baseline SI benefited most from CKM (Pearson \(r=-0.653\),
\(p=0.002\), 19 models; Figure 5). CKM
narrowed the SI range from 0.907--0.970 (Control) to 0.927--0.976 (CKM),
acting as a behavioral-variance reducer whose effect was inversely
proportional to intrinsic consistency (Figure 5). Family-level analysis confirmed
differential responsiveness (Kruskal-Wallis \(H=13.57\), \(p=0.004\)):
GPT (+0.031) \textgreater{} Gemini (+0.016) \textgreater{} Grok (+0.007)
\textgreater{} Claude (+0.001).

\includegraphics{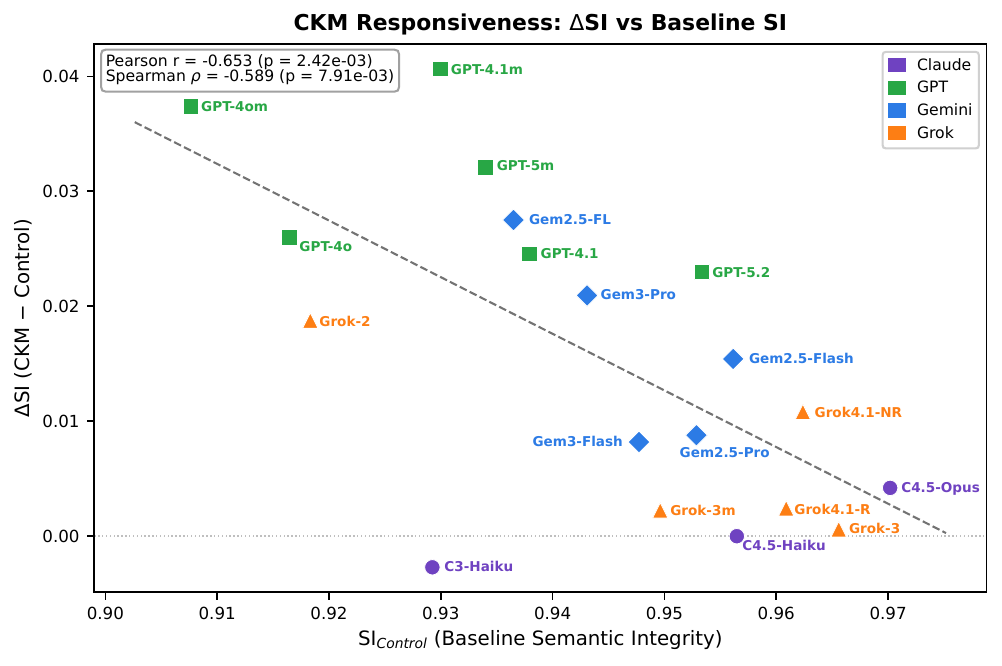}

\textbf{Figure 5. Behavioral-variance reducer effect.} Each point
represents one of the 19 Legacy models; the x-axis shows baseline SI under Control (axis truncated, does not start at 0), the
y-axis shows CKM-induced \(\Delta\)SI. The negative correlation
(\(r=-0.653\), \(p=0.002\)) indicates that models with lower baseline
consistency show larger CKM effects.

\hypertarget{scenario-level-variation}{%
\paragraph{Scenario-Level Variation}\label{scenario-level-variation}}

The effect size depends not only on which model is used but on what
scenario type the model faces.

CKM effectiveness varied with task characteristics. In the NewModels
cohort, ambiguity-dominant scenarios (S1, S4, S6) showed mean
\(\Delta\)SI of +0.0416, compared to +0.0443 for non-ambiguity scenarios
(S2, S3, S5, S9, S10), indicating comparable magnitude across scenario
types. The Legacy cohort showed a more pronounced differentiation
(ambiguity +0.046 vs.~non-ambiguity +0.019, 2.4-fold difference, see
Supplementary §S7.2), suggesting potential cohort-specific mechanisms.
The convergence across scenario types in NewModels may indicate that in
more recent models, CKM's benefit derives less from ambiguity-specific
epistemic resolution and more from general structural clarity.

This pattern is consistent with the hypothesis that CKM operates through
explicit knowledge categorization (see §7.2 for mechanism discussion),
though the relative importance of ambiguity resolution versus
format-driven clarity may differ across model architectures and training
paradigms. Extended scenario-level analysis is provided in Supplementary
§S7 and §S-Profiles.

\hypertarget{robustness-checks}{%
\subsubsection{6.4 Robustness Checks}\label{robustness-checks}}

Before interpreting the preceding results, we address potential
confounds: data integrity, embedding backbone sensitivity, and
temperature effects.

We verified data integrity across all experiments. For Exp1--3 and Study
A/B, the LLM-only valid rate was 98.4\% (24,054/24,443); including the
Exp1 human pilot, the rate is 96.7\% (24,092/24,923). Exp6 sham-arm
arm-level extraction rates range 93.8--99.8\% (lowest individual cell 78.75\%, gemini-2.5-pro Sham-B; see
Supplementary §S-Exp6.4 for per-model detail). Canonical datasets were
verified via SHA-256 checksums (Supplementary §S2, §S-Ablation.4 for
per-experiment breakdown).

Embedding sensitivity analysis showed that mpnet detected significant
CKM effects while MiniLM yielded null results (\(d=0.05\), \(p=0.844\))
due to cosine ceiling compression on Korean text, confirming that
embedding backbone selection is a critical methodological decision
(Supplementary §S1).

Temperature robustness was assessed at \(T=0.7\) (\(N=728\)). All arms
showed SI degradation under higher temperature, but the magnitude of
degradation was monotonically ordered by structural complexity:

\textbf{Table 9. SI drop by arm (\(T=0\) → \(T=0.7\)).} Mean SI under
deterministic (\(T=0\)) and stochastic (\(T=0.7\)) sampling, averaged
across the temperature-probe subcohort (\(N=728\)). The CKM arm halves
the temperature-induced consistency penalty relative to Control.

\begin{longtable}[]{@{}llll@{}}
\toprule\noalign{}
Arm & SI (\(T=0\)) & SI (\(T=0.7\)) & Drop \\
\midrule\noalign{}
\endhead
\bottomrule\noalign{}
\endlastfoot
Control & 0.954 & 0.913 & \textbf{−0.041} \\
JSON-Only & 0.967 & 0.941 & −0.026 \\
F/H/E-Only & 0.981 & 0.959 & −0.022 \\
Full CKM & 0.984 & 0.965 & \textbf{−0.019} \\
\end{longtable}

CKM halved the consistency penalty from sampling noise (2.0\(\times\)
ratio) (Table 9). Previously marginal effects (H2: \(g=0.99\) \(\rightarrow\)
2.87) reached significance under higher temperature, driven by
disproportionate Control degradation rather than CKM improvement. CKM
thus functions as a stochasticity buffer (full results in Supplementary
§S-Ablation).

\hypertarget{sham-restriction-ablation-exp6}{%
\subsubsection{6.5 Sham-Restriction Ablation
(Exp6)}\label{sham-restriction-ablation-exp6}}

The 4-arm ablation (§6.1) established that F/H/E semantic separation is
the primary mechanism, and §7.2 argued against output-space compression
using SI\(\times\)DFR orthogonality. However, the 4-arm design did not
include a sham-restriction control that matches CKM in structural
complexity without epistemic content. Exp6 addresses this gap with a
5-arm design that directly decomposes the CKM effect into structural
scaffolding versus reasoning-semantic contributions.

\hypertarget{design}{%
\paragraph{Design}\label{design}}

Five arms were tested using the same 12 primary NewModels \(\times\) 8
scenarios \(\times\) \(K=10\) protocol (\(N=3\),840 per arm):

\begin{itemize}
\tightlist
\item
  \textbf{Vanilla}: No structure (Control baseline, reused from
  Exp2-NM).
\item
  \textbf{Sham-A} (Surface Semantic): JSON schema with non-reasoning
  semantic slot names (color, animal, weather, food, mood) matched to
  CKM in slot count and format complexity.
\item
  \textbf{Sham-B} (Non-Reasoning Semantic): JSON schema with contextual
  but non-epistemic slot names (time, place, participant, activity,
  object).
\item
  \textbf{Sham-C} (Symbolic/Meaningless): JSON schema with arbitrary
  symbolic slot names (V, W, X, \(\alpha\), \(\beta\), slot-1)
  containing no semantic content.
\item
  \textbf{CKM}: F/H/E epistemic role separation with full state
  enforcement (reused from Exp3-NM).
\end{itemize}

This design isolates three components: (1) structural scaffolding (any
JSON schema vs.~no schema), (2) non-epistemic semantic content
(meaningful but non-reasoning slot names vs.~symbolic), and (3)
reasoning-semantic content (F/H/E epistemic categories vs.~non-epistemic
alternatives). All three sham arms match CKM in format complexity, slot
count, and approximate token overhead, controlling for the structural
compression confound.

\hypertarget{aggregate-results}{%
\paragraph{Aggregate Results}\label{aggregate-results}}

\textbf{Table 10. 5-arm aggregate metrics (primary 12-model basis).}

\begin{longtable}[]{@{}lllll@{}}
\toprule\noalign{}
Arm & SI & DFR\(_{\text{persist}}\) & SDR & FCS \\
\midrule\noalign{}
\endhead
\bottomrule\noalign{}
\endlastfoot
Vanilla & 0.952 & 0.080 & --- & --- \\
Sham-A & 0.962 & 0.082 & 0.920 & --- \\
Sham-B & 0.962 & 0.419 & 0.890 & --- \\
Sham-C & 0.962 & 0.116 & 0.928 & --- \\
\textbf{CKM} & \textbf{0.974} & \textbf{0.069}\(^\|\) & \textbf{0.939} &
--- \\
\end{longtable}

\(^\|\) CKM DFR\(_{\text{persist}}\) is from Exp3 CKM\_PERSIST condition
(\passthrough{\lstinline!exp3\_newmodels\_metrics\_v1.3\_active12.json!});
all five arms are aligned to the persist protocol (single-turn for
Vanilla / Sham-A/B/C; CKM\_PERSIST for the CKM arm). The 5-arm SI values
are reported on the raw-output substrate to maintain comparability with
the sham arms (which lack a parsed-field representation by design); the
analogous value-only sensitivity recomputation for Vanilla and CKM
yields the same SI ordering with magnitudes reduced toward the §6.1
value-only baseline (see Supplementary §S-Robustness for the
value-only methodology). Table 10 was regenerated from
\passthrough{\lstinline!exp6\_5way\_comparison\_v1.2\_active12.json!}
(CKM column corrected from a prior source-field misalignment in the v1.1
canonical; see §S-Exp6.5).

The SI ordering is: Vanilla (0.9521) \textless{} Sham-A (0.9616)
\(\approx\) Sham-B (0.9623) \(\approx\) Sham-C (0.9617) \(\ll\) CKM
(0.9741). All three sham arms cluster at a similar SI level,
approximately 1 percentage point above Vanilla but 1.2 percentage points
below CKM. The near-equivalence among sham arms (pairwise
\(|g| < 0.07\), all \(p_{\text{Holm}} = 1.0\)) indicates that the
specific semantic content of slot names does not materially influence
SI; what matters is whether the slots enforce epistemic role separation.

The DFR\(_{\text{persist}}\) ordering differs sharply: CKM (0.069)
\textless{} Vanilla (0.080) \(\approx\) Sham-A (0.082) \textless{}
Sham-C (0.116) \(\ll\) Sham-B (0.419). CKM is the only arm that reduces
DFR below Vanilla. Sham-B shows a dramatic 5.2-fold increase relative to
Vanilla.

\textbf{Table 11. Pairwise Hedges' \(g\) for SI (upper triangle) with
Holm-corrected \(p\)-values (12 primary models, 10 pairwise
comparisons).} Entries are \(g_{ij}\) = (row mean − column mean) /
pooled SD. CKM-vs-Vanilla is the strongest SI contrast; CKM exceeds
each sham arm individually, while the three sham arms are pairwise
indistinguishable on SI.

\begin{longtable}[]{@{}
  >{\raggedright\arraybackslash}p{(\columnwidth - 10\tabcolsep) * \real{0.0714}}
  >{\raggedright\arraybackslash}p{(\columnwidth - 10\tabcolsep) * \real{0.0714}}
  >{\raggedright\arraybackslash}p{(\columnwidth - 10\tabcolsep) * \real{0.2041}}
  >{\raggedright\arraybackslash}p{(\columnwidth - 10\tabcolsep) * \real{0.2143}}
  >{\raggedright\arraybackslash}p{(\columnwidth - 10\tabcolsep) * \real{0.2041}}
  >{\raggedright\arraybackslash}p{(\columnwidth - 10\tabcolsep) * \real{0.2347}}@{}}
\toprule\noalign{}
\begin{minipage}[b]{\linewidth}\raggedright
\end{minipage} & \begin{minipage}[b]{\linewidth}\raggedright
Vanilla
\end{minipage} & \begin{minipage}[b]{\linewidth}\raggedright
Sham-A
\end{minipage} & \begin{minipage}[b]{\linewidth}\raggedright
Sham-B
\end{minipage} & \begin{minipage}[b]{\linewidth}\raggedright
Sham-C
\end{minipage} & \begin{minipage}[b]{\linewidth}\raggedright
CKM
\end{minipage} \\
\midrule\noalign{}
\endhead
\bottomrule\noalign{}
\endlastfoot
\textbf{Vanilla} & --- & \(-0.91\) (\(p\) = .094) & \(-1.14\) (\(p\) =
.015*) & \(-0.74\) (\(p\) = .166) & \(-2.71\) (\(p\) \textless{}
.001***) \\
\textbf{Sham-A} & & --- & \(-0.07\) (\(p=1.0\)) & \(-0.01\) (\(p=1.0\))
& \(-1.23\) (\(p\) = .015*) \\
\textbf{Sham-B} & & & --- & \(+0.05\) (\(p=1.0\)) & \(-1.36\) (\(p\) =
.008**) \\
\textbf{Sham-C} & & & & --- & \(-0.97\) (\(p\) = .095) \\
\textbf{CKM} & & & & & --- \\
\end{longtable}

\(p\)-values are Holm-Bonferroni corrected across the 10 pairwise SI
comparisons per metric (canonical:
\passthrough{\lstinline!exp6\_5way\_comparison\_v1.2\_active12.json!}
\passthrough{\lstinline!SI.p\_holm\_matrix!}). Hedges' \(g\) values are
reported using the canonical independent-groups convention (n = 12 per
arm); paired-design effect sizes (which would be larger by a factor of
approximately 2 for Vanilla-vs-CKM, given the within-subject design) are
provided as a sensitivity in Supplementary §S-Exp6 alongside this
primary table.

CKM significantly exceeds Sham-A and Sham-B on SI (\(g=1.23\)--1.36,
\(p_{\text{Holm}} < .05\)) and shows a large effect versus Sham-C
(\(g=0.97\), \(p_{\text{Holm}}\) = .095). The three sham arms are
mutually indistinguishable (\(|g| < 0.07\), all
\(p_{\text{Holm}} = 1.0\)). This pattern directly contradicts the
compression hypothesis: if CKM's SI gain were attributable to structural
restriction alone, the sham arms (which impose equivalent structural
restriction with non-epistemic content) would match CKM. They do not.

\hypertarget{si-decomposition}{%
\paragraph{SI Decomposition}\label{si-decomposition}}

The 5-arm hierarchy enables a quantitative decomposition of the CKM SI
improvement (\(\Delta\)SI = +0.022 relative to Vanilla):

\textbf{Table 12. SI decomposition: structural scaffolding vs.~F/H/E
semantic content.} The Vanilla → CKM SI gain is partitioned via the
sham hierarchy. Structural scaffolding (any JSON schema with named
slots) accounts for \textasciitilde{}45\% of the effect; F/H/E
epistemic role separation contributes the remaining \textasciitilde{}55\%.

\begin{longtable}[]{@{}
  >{\raggedright\arraybackslash}p{(\columnwidth - 6\tabcolsep) * \real{0.3421}}
  >{\raggedright\arraybackslash}p{(\columnwidth - 6\tabcolsep) * \real{0.3947}}
  >{\raggedright\arraybackslash}p{(\columnwidth - 6\tabcolsep) * \real{0.1316}}
  >{\raggedright\arraybackslash}p{(\columnwidth - 6\tabcolsep) * \real{0.1316}}@{}}
\toprule\noalign{}
\begin{minipage}[b]{\linewidth}\raggedright
Component
\end{minipage} & \begin{minipage}[b]{\linewidth}\raggedright
Transition
\end{minipage} & \begin{minipage}[b]{\linewidth}\raggedright
\(\Delta\)SI
\end{minipage} & \begin{minipage}[b]{\linewidth}\raggedright
Proportion
\end{minipage} \\
\midrule\noalign{}
\endhead
\bottomrule\noalign{}
\endlastfoot
Structural scaffold & Vanilla \(\rightarrow\) Sham-C & +0.010 & 45\% \\
Reasoning-semantic (F/H/E) & Sham cluster \(\rightarrow\) CKM & +0.012 &
55\% \\
\textbf{Total} & Vanilla \(\rightarrow\) CKM & \textbf{+0.022} &
\textbf{100\%} \\
\end{longtable}

Surface semantic content (Sham-A vs.~Sham-C: \(\Delta\)SI \(\approx\)
0.000) and non-reasoning semantic content (Sham-B vs.~Sham-A:
\(\Delta\)SI \(\approx\) 0.000) contribute negligibly. The entire
sham-attributable SI gain comes from structural scaffolding (the mere
presence of a JSON schema with named slots), while the majority of the
CKM effect (55\%) derives from the specific epistemic content of F/H/E
decomposition (Table 12). This confirms that CKM's consistency improvement is not
reducible to structural restriction.

\hypertarget{si-dfr-dissociation}{%
\paragraph{SI-DFR Dissociation}\label{si-dfr-dissociation}}

The most striking finding is the dissociation between SI and DFR across
the five arms:

\textbf{Table 13. Per-arm joint SI/DFR\(_{\text{persist}}\) improvement
relative to Vanilla.} Only CKM achieves simultaneous SI gain and DFR
reduction; all three sham arms improve SI but degrade DFR, falsifying
the pure compression hypothesis.

\begin{longtable}[]{@{}
  >{\raggedright\arraybackslash}p{(\columnwidth - 6\tabcolsep) * \real{0.0822}}
  >{\raggedright\arraybackslash}p{(\columnwidth - 6\tabcolsep) * \real{0.3288}}
  >{\raggedright\arraybackslash}p{(\columnwidth - 6\tabcolsep) * \real{0.3425}}
  >{\raggedright\arraybackslash}p{(\columnwidth - 6\tabcolsep) * \real{0.2466}}@{}}
\toprule\noalign{}
\begin{minipage}[b]{\linewidth}\raggedright
Arm
\end{minipage} & \begin{minipage}[b]{\linewidth}\raggedright
\(\Delta\)SI (vs.~Vanilla)
\end{minipage} & \begin{minipage}[b]{\linewidth}\raggedright
\(\Delta\)DFR (vs.~Vanilla)
\end{minipage} & \begin{minipage}[b]{\linewidth}\raggedright
Joint improvement?
\end{minipage} \\
\midrule\noalign{}
\endhead
\bottomrule\noalign{}
\endlastfoot
Sham-A & +0.010 & +0.002 (marginal) & No \\
Sham-B & +0.010 & +0.339 (much worse) & No \\
Sham-C & +0.010 & +0.036 (worse) & No \\
\textbf{CKM} & \textbf{+0.022} & \textbf{\(-0.011\) (better)} &
\textbf{Yes} \\
\end{longtable}

CKM is the only arm that achieves simultaneous SI improvement and DFR
reduction (Figure 6). All three sham arms improve SI but degrade DFR,
demonstrating that structural scaffolding without epistemic role
separation produces a consistency-stability tradeoff rather than joint
improvement (Table 13). This dissociation constitutes direct evidence against the
pure compression hypothesis: under compression, SI\(\uparrow\) should
entail DFR\(\downarrow\) (a narrower output space contains fewer
distinct actions), but the sham arms show the opposite pattern.

\includegraphics{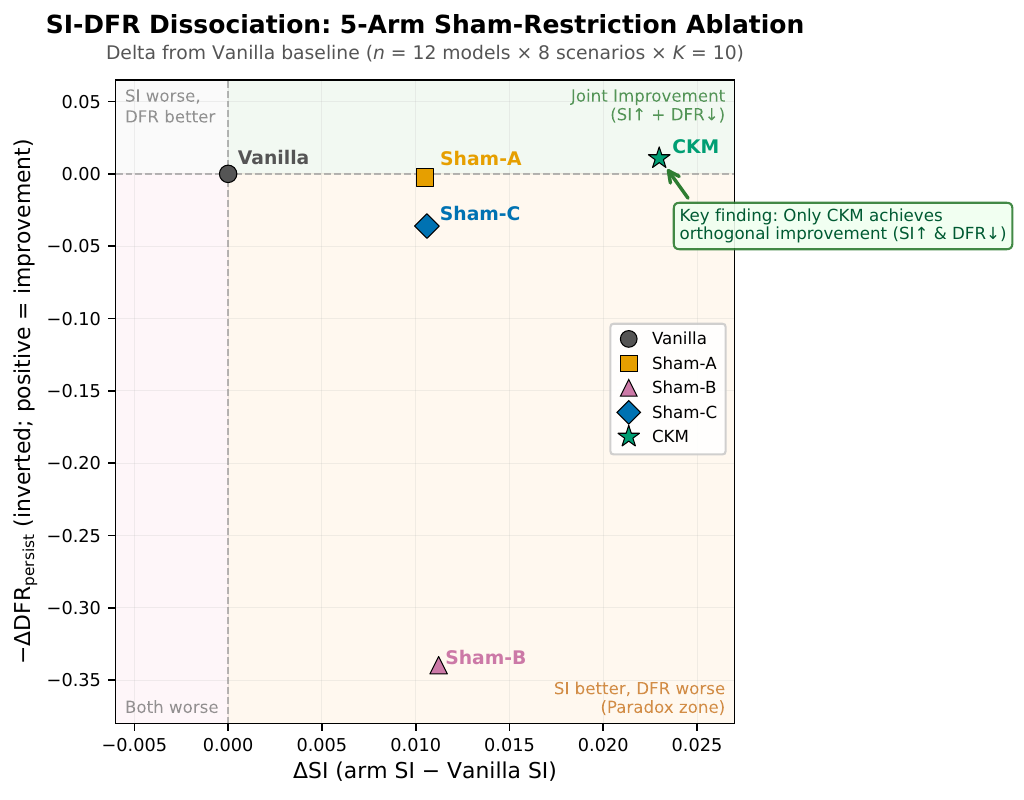}

\textbf{Figure 6. SI-DFR dissociation scatter for the 5-arm
sham-restriction ablation.} X-axis: \(\Delta\)SI relative to Vanilla
(positive = higher consistency). Y-axis:
\(-\Delta\)DFR\(_{\text{persist}}\) relative to Vanilla (positive = more
stable). Vanilla is plotted at the origin. The upper-right quadrant
represents joint improvement (SI\(\uparrow\) + DFR\(\downarrow\)); only
CKM occupies this region. Sham-B (lower-right) exhibits the paradox of
improved SI but dramatically worsened DFR. Sham-A and Sham-C show modest
SI gains with slight DFR degradation.

\hypertarget{sham-b-paradox}{%
\paragraph{Sham-B Paradox}\label{sham-b-paradox}}

Sham-B presents a noteworthy anomaly: it achieves SI comparable to the
other sham arms (0.962) yet exhibits DFR\(_{\text{persist}}\) = 0.419,
over five times the Vanilla rate (0.080). We hypothesize that
non-reasoning semantic context (time, place, participant) disrupts
action stability under state injection because these contextual framings
are semantically rich enough to trigger re-evaluation but lack the
epistemic structure to anchor decisions. When a model encounters
persisted state containing ``morning/school/self'' contextual labels, it
may treat these as novel situational cues warranting decision revision,
rather than as authoritative context to be preserved.

By contrast, CKM's F/H/E labels explicitly tag information by epistemic
provenance, providing an anchoring structure that resists
state-injection-induced re-evaluation. This account remains a hypothesis
(claim strength: below \(\beta\)) pending targeted experiments that
manipulate contextual framing independently of slot count.

\hypertarget{judgment-r4-mechanism-decomposition-claim-beta}{%
\paragraph{\texorpdfstring{Judgment: R4 (Mechanism Decomposition), Claim
\(\beta\)}{Judgment: R4 (Mechanism Decomposition), Claim \textbackslash beta}}\label{judgment-r4-mechanism-decomposition-claim-beta}}

The 5-arm results are most consistent with the R4 (mechanism
decomposition) scenario: CKM effects decompose into structural
scaffolding (\textasciitilde45\%) and reasoning-semantic-specific
(\textasciitilde55\%) components, with the two components contributing
to different consistency dimensions (structural scaffolding to SI only;
reasoning-semantic to both SI and DFR). This decomposition meets the
\(\beta\) (medium) claim threshold: the effect is replicated across 12
primary models, the pairwise contrasts are statistically significant
after Holm correction for the critical comparisons (CKM vs.~Sham-A/B:
\(p < .046\); CKM vs.~Vanilla: \(p < .001\)), and the SI-DFR
dissociation provides a mechanistic signature that is not predicted by
simpler accounts.

\hypertarget{case-study-behavioral-profile-of-a-compromised-model-gpt-4.1-mini}{%
\subsubsection{6.6 Case Study: Behavioral Profile of a Compromised Model
(gpt-4.1-mini)}\label{case-study-behavioral-profile-of-a-compromised-model-gpt-4.1-mini}}

This section presents a case study of gpt-4.1-mini, the sole model
excluded from the primary NewModels analysis. The purpose is twofold: to
provide full transparency about the exclusion decision and to document a
behaviorally anomalous profile that may be of independent scientific
interest.

\hypertarget{background}{%
\paragraph{Background}\label{background}}

OpenAI issued a deprecation notice for gpt-4.1-mini on 2026-02-13, two
days before our data collection began (2026-02-15). During the
collection window, independent reports documented degradation in
instruction-following behavior. These circumstances raise concerns about
whether gpt-4.1-mini's behavioral data reflect stable model
characteristics or transient degradation artifacts.

\hypertarget{behavioral-profile}{%
\paragraph{Behavioral Profile}\label{behavioral-profile}}

Under the full 13-model cohort, gpt-4.1-mini exhibited a distinctive
behavioral signature. In Exp3 (state persistence), gpt-4.1-mini was the
sole model where DFR\_persist (0.281) exceeded DFR\_ckm (0.231),
indicating that state injection increased rather than decreased decision
instability. This reversal was unique among all 13 models.

In Exp6 (5-arm sham-restriction ablation), gpt-4.1-mini showed
dramatically elevated DFR\_persist across all sham arms. The 5-arm DFR
profile (all-13 values) is:

\textbf{Table 14. gpt-4.1-mini DFR\(_{\text{persist}}\) across Exp6 arms
(all-13 basis).}

\begin{longtable}[]{@{}ll@{}}
\toprule\noalign{}
Arm & DFR\(_{\text{persist}}\) \\
\midrule\noalign{}
\endhead
\bottomrule\noalign{}
\endlastfoot
Vanilla & 0.679 \\
Sham-A & 0.912 \\
Sham-B & 0.900 \\
Sham-C & 0.888 \\
CKM & 0.281 \\
\end{longtable}

CKM reduced gpt-4.1-mini's DFR from the Vanilla rate of 0.679 to 0.281,
a 59\% reduction. This suggests that F/H/E epistemic role separation
partially stabilized even a compromised model. However, the residual DFR
of 0.281 under CKM is substantially higher than the 12-model primary
median of 0.000, and the extreme sensitivity to all sham arms (DFR
\textgreater{} 0.88) is consistent with a model in a degraded
operational state.

\hypertarget{interpretation}{%
\paragraph{Interpretation}\label{interpretation}}

We hypothesize that gpt-4.1-mini's behavior during the collection window
reflected transient instability associated with the deprecation process,
potentially including weight modifications, routing changes, or reduced
infrastructure prioritization. Under this hypothesis, the model's
elevated DFR across all conditions reflects a global instability mode
rather than a condition-specific interaction. The partial stabilization
under CKM is consistent with the general finding that F/H/E epistemic
structure provides an anchoring effect, but the compromised baseline
limits the interpretive value of condition-specific comparisons.

\hypertarget{limitations}{%
\paragraph{Limitations}\label{limitations}}

This case study has several limitations. First, \(N=1\): the behavioral
profile describes a single model under specific temporal circumstances
and cannot be generalized. Second, we lack vendor documentation of what
changes, if any, were applied to gpt-4.1-mini during the deprecation
window. Third, we cannot distinguish between model-intrinsic instability
and infrastructure-level variability.

\hypertarget{exclusion-justification}{%
\paragraph{Exclusion Justification}\label{exclusion-justification}}

The exclusion of gpt-4.1-mini from primary analyses satisfies five
criteria:

\begin{enumerate}
\def\labelenumi{\arabic{enumi}.}
\tightlist
\item
  \textbf{Vendor-issued deprecation}: OpenAI published a deprecation
  notice on 2026-02-13, overlapping with our collection window.
\item
  \textbf{Independent degradation reports}: External users documented
  instruction-following degradation during this period.
\item
  \textbf{Anomalous behavioral profile}: gpt-4.1-mini was the sole model
  where state injection increased DFR (DFR\_persist \textgreater{}
  DFR\_ckm), a pattern not observed in any other model.
\item
  \textbf{Directional preservation}: All primary findings (SI
  improvement, DFR reduction, mechanism decomposition) are preserved
  under both the 12-model and 13-model cohorts (Supplementary
  §S-Sensitivity).
\item
  \textbf{Full transparency}: Complete data are reported in this case
  study and the sensitivity analysis; no data are suppressed.
\end{enumerate}

\begin{center}\rule{0.5\linewidth}{0.5pt}\end{center}

\hypertarget{discussion}{%
\subsection{7. Discussion}\label{discussion}}

The results above support a narrow but deployment-relevant claim: CKM
increases behavioral consistency in the tested LLM decision scenarios,
and the effect is not reducible to JSON formatting alone. This section
interprets that claim without expanding it into stronger claims about
correctness, ethics, or general intelligence. The discussion therefore
follows three boundaries: what was measured, what the ablations support
mechanistically, and what remains outside the present evidence.

The empirical message of §6 is that behavioral consistency is
measurable, model-dependent, and partially reducible through F/H/E state
enforcement. CKM improves repeated-output stability in most tested
models, but the effect is not a generic consequence of making the output
more structured. The ablations show that the semantic role of the slots
matters: Fact, Heuristic, and Emotion separation produces a different
behavioral signature from non-epistemic JSON scaffolds.

This section interprets those findings. §7.1 argues that behavioral
consistency should be treated as a candidate evaluation dimension
alongside capability. §7.2 explains why structured state enforcement may
work, while separating confirmed findings from hypotheses. §7.3
discusses the human pilot. §7.4 outlines practical implications of
state-persistence variation. §7.5 states the scope limits, and §7.6
identifies the most important next experiments.

\hypertarget{behavioral-consistency-as-an-evaluation-dimension}{%
\subsubsection{7.1 Behavioral Consistency as an Evaluation
Dimension}\label{behavioral-consistency-as-an-evaluation-dimension}}

The results presented in §6 establish six empirical findings; we now
consider their broader implications, beginning with what they reveal
about consistency as an evaluation dimension. Our data suggest that
behavioral consistency is distinct from the dimensions captured by
current capability benchmarks: this study did not measure capability
scores directly, but the within-suite dissociations we observe (e.g.,
models with high SI but high DFR, or high FCS but low CRR) imply that no
single capability proxy in our suite predicts the others.

The Tier B phenomenon, in which models achieve high single-turn
consistency yet poor multi-turn stability, illustrates a failure mode
that is not captured by any of our single-turn measurements alone. The
metric suite captures failure modes that are partially independent of
one another, and we treat the relationship between behavioral
consistency and external capability benchmarks as an open empirical
question for joint-measurement studies. Consistency profiling may
therefore merit consideration as a candidate complement to existing LLM
evaluation dimensions, pending such joint validation.

\hypertarget{why-structured-state-enforcement-works}{%
\subsubsection{7.2 Why Structured State Enforcement
Works}\label{why-structured-state-enforcement-works}}

Given that CKM improves consistency in our experiments, a natural
question is why it works. Before proposing mechanistic accounts, we
first address a prior concern: whether the observed consistency gains
could be explained trivially by output-space compression, without
invoking any reorganization of reasoning. We then evaluate four
candidate mechanisms as an interpretive framework consistent with the
empirical patterns; the value-only ablation (§6.1) rules out one of them
as a measurement artifact, leaving three operative mechanisms (cognitive
load reduction, epistemic boundary enforcement, and structural
scaffolding as enabling context).

Each of the three receives varying degrees of direct support from our
data, and we note where the evidence is indirect rather than causal.

\textbf{Disentangling Restriction from Organization.}

A natural concern is that any form of output restriction would trivially
reduce variability, rendering the observed consistency gains
uninformative. Under this view, CKM's F/H/E scaffold would function as a
more elaborate form of response-space compression rather than as
reasoning organization, and the improved Semantic Integrity would
reflect narrower permissible outputs rather than better-structured
reasoning.

We address this concern on two levels. First, the value-only SI analysis
(§6.1, Supplementary §S-Robustness) directly controls for structural
homogeneity: when JSON surface features are removed from the SI
computation, JSON-only structural enforcement shows no significant
effect on semantic consistency (\(g=-0.20\), non-significant), while FHE
semantic separation preserves its effect (\(g=2.24\)). Structural
scaffolding alone is insufficient; the operative component is epistemic
categorization.

Second, and more decisively, the relationship between single-turn
semantic consistency (SI) and multi-turn decision stability (DFR)
refutes the uniform-compression account. Under a pure compression
mechanism, both axes should co-move: any restriction that tightens SI by
narrowing the response space should also tighten DFR, because a narrower
space contains fewer distinct actions. The data do not exhibit this
co-movement. Across the 24-model union pool (23 models with paired SI
and CKM\_PERSIST DFR measurements), SI and DFR do not co-move positively
as compression would predict; if anything the association is weakly
negative (Pearson \(r=-0.18\), \(p=0.41\); Spearman \(\rho=-0.40\),
\(p=0.06\)).

When the SI\(\times\)DFR plane is partitioned by median-split (SI median
= 0.967, DFR cutoff = 0.15), models populate all four quadrants:
upper-right (high-SI, high-DFR) \(n=2\); upper-left (high-SI, low-DFR)
\(n=9\); lower-right (low-SI, high-DFR) \(n=5\); lower-left (low-SI,
low-DFR) \(n=7.\) The upper-right quadrant (high SI, high CKM\_PERSIST
DFR) is occupied by gemini-2.5-flash and gpt-4.1-mini, which achieve SI
above the median under CKM while exhibiting substantial decision flips
under state injection (gemini-2.5-flash DFR\_persist 0.82); gemini-2.5-pro
(DFR\_persist 0.644) sits just below the SI median in the lower-right
region.

A uniform output-space compression would predict an empty upper-right
quadrant; its observed non-emptiness with frontier occupants refutes the
prediction. This median-split lens is complementary to the fixed-cutoff
Tier taxonomy of §6.3; both partition the same plane but under distinct
cutoff schemes.

We acknowledge that orthogonality between axes does not, by itself,
causally prove that F/H/E organizes reasoning rather than compressing
output space along a different axis. The 5-arm sham-restriction ablation
(§6.5) provides a direct empirical test: three sham arms matched to CKM
in format complexity and slot count but containing non-epistemic content
all failed to replicate CKM's joint SI\(\uparrow\) + DFR\(\downarrow\)
pattern, confirming that structural restriction alone is insufficient.

\textbf{Exp6 Mechanism Decomposition.}

The 5-arm sham-restriction ablation (§6.5) resolves the outstanding
question from the preceding analysis by providing a direct causal test
of whether F/H/E semantic separation organizes reasoning or merely
compresses output space along a different axis.

The compression hypothesis predicted that sham arms, which impose
equivalent structural restriction with non-epistemic content, would
match CKM on all consistency axes. The organization hypothesis predicted
that sham arms and CKM would converge on SI (both involve structural
scaffolding) but diverge on DFR (only CKM provides epistemic anchoring).
The data strongly favor the organization account: all three sham arms
achieve SI approximately 1 percentage point above Vanilla (structural
scaffold effect, \textasciitilde45\% of the total CKM gain), but none
consistently reduces DFR below Vanilla.

CKM alone achieves the remaining 55\% of the SI gain and is the only arm
that simultaneously reduces DFR, a pattern consistent with reasoning
reorganization rather than output-space narrowing.

The SI decomposition (structural scaffold \textasciitilde45\%,
reasoning-semantic \textasciitilde55\%) provides a quantitative
partition of the CKM effect that was previously available only as a
qualitative argument from the 4-arm ablation. The structural component
corresponds to the format enforcement effect identified in §6.1, while
the reasoning-semantic component corresponds to the F/H/E semantic
separation effect. Importantly, the sham arms demonstrate that the
structural component contributes to SI but not to DFR, while the
reasoning-semantic component contributes to both, establishing that
these are functionally distinct mechanisms rather than different aspects
of the same process.

\textbf{Sham-B Paradox and Contextual Disruption (Hypothesis).}

The Sham-B arm presents a theoretically informative anomaly. Despite
achieving SI comparable to the other sham arms, Sham-B exhibits
DFR\(_{\text{persist}}\) = 0.419, over five times the Vanilla rate. We
hypothesize that non-reasoning semantic context (time, place,
participant) disrupts action stability under state injection because
these contextual framings are semantically rich enough to trigger
re-evaluation but lack the epistemic structure to anchor decisions. This
hypothesis is consistent with the observation that 11/12 primary models
show higher DFR under Sham-B than Sham-A, and 11/12 show higher DFR
under Sham-B than Vanilla.

If confirmed, this would suggest that the epistemic anchoring provided
by F/H/E is not merely additive but protective: it prevents semantically
rich context from destabilizing decisions. This account remains a
hypothesis pending targeted experiments that manipulate contextual
framing independently of slot count; we do not elevate it to a confirmed
finding.

\textbf{Mechanism Decomposition.}

The first mechanism is cognitive load reduction through forced
categorization. CKM imposes an epistemic hygiene discipline, requiring
explicit labeling of observable facts, inferred heuristics, and
affective signals before reasoning proceeds, thereby reducing the space
of possible interpretations. Without this separation, the model must
simultaneously determine what information means and what kind of
information it is; with separation, the categorization is resolved at
the input stage, freeing subsequent processing to operate on
pre-classified inputs. This mechanism predicts that CKM should be most
effective in ambiguity-dominant scenarios, where fact-inference
conflation is greatest.

The scenario-level data are consistent with this prediction in the
Legacy cohort (§6.3): ambiguity scenarios show 2.4-fold larger
\(\Delta\)SI (+0.046 vs.~+0.019) than scenarios with clearer factual
constraints. In the primary NewModels cohort, however, this
differentiation is largely absent (ambiguity +0.0416 vs.~non-ambiguity
+0.0443, 1.07-fold ratio), suggesting that newer models may distribute
the F/H/E benefit more uniformly across scenario types; this
cohort-dependent pattern warrants targeted investigation rather than
serving as standalone confirmation of the cognitive-load mechanism.

The second mechanism is epistemic boundary enforcement. The mandatory
distinction between \(F\) and \(H\), with explicit uncertainty markers
on heuristic content, makes the epistemic status of each reasoning
component transparent. This can reduce a common failure mode in which
models present speculative inferences with the same confidence as
verified observations, a conflation that increases the variance of
downstream conclusions. The human pilot data (§7.3) provide
complementary evidence: participants achieved perfect F/H/E compliance
across the manually reviewed subset (0 violations in 15 annotated CKM
responses; see Supplementary §S11.2 for the review protocol), suggesting
that the boundaries correspond to cognitively natural categorization
rather than arbitrary imposition.

The third mechanism is output space reduction through structural
scaffolding. The requirement that each response populate specific fields
(facts, heuristics, emotion code, action, reason) constrains the
response space. The original SI measurement attributed a large effect to
this mechanism (\(g=1.87\)), but value-only analysis (§6.1,
Supplementary §S-Robustness) revealed that this measured effect
primarily reflects JSON structural homogeneity inflating cosine
similarity. The format enforcement component may still contribute to
behavioral consistency through channels not captured by SI---such as
ensuring complete slot coverage or reducing length variability---but its
contribution to semantic convergence is not separable from measurement
artifact within the current framework.

This finding elevates the second mechanism (epistemic boundary
enforcement) and the first (cognitive load reduction through forced
categorization) as the primary explanatory accounts, both of which
operate through F/H/E semantic separation rather than format
constraints.

The fourth consideration is the relationship between format and semantic
constraints. The original SI measurement suggested super-additive
synergy (\(g=0.835\), \(p=0.0199\)), but the value-only analysis indicates
that the apparent synergy was driven by JSON structural homogeneity
compounding with semantic convergence in the Full CKM condition. After
artifact removal, the residual advantage of Full CKM over FHE-Only is
not attributable to genuine semantic synergy. We therefore consolidate
the mechanistic account around three operative mechanisms (cognitive
load reduction, epistemic boundary enforcement, and structural
scaffolding as enabling context), with semantic separation as the
primary active ingredient.

This consolidated interpretation is offered as a theoretical framework
consistent with the corrected data; direct causal evidence would require
interventions that independently manipulate each mechanism, which we
identify as a direction for future work.

It is important to note, however, that reduced variability does not
imply improved reasoning quality. Higher SI indicates convergence, but
convergence toward shared error remains possible. Our framework measures
consistency, not accuracy, and these dimensions require independent
assessment.

\hypertarget{human-baseline-comparison}{%
\subsubsection{7.3 Human Baseline
Comparison}\label{human-baseline-comparison}}

While the preceding mechanisms were developed to explain LLM behavior,
the human pilot provides preliminary evidence that similar dynamics
operate in human reasoning. Exp1's small sample (\(n=6\)) precludes
statistical generalization, but three directional findings merit
attention. CKM surfaced suppressed affective states (anxiety reports: 0
→ 6; ``no emotion'': 5 → 2), suggesting that structured prompting
reorganizes the reasoning process rather than merely constraining it.

F/H/E separation compliance was perfect (0 violations across the 15 CKM
responses for which the pilot protocol included manual compliance
annotations; 4 additional CKM responses lacked reviewer annotations and
were excluded from this subset; see Supplementary §S11.2), indicating
that the structural requirements are cognitively natural for human
reasoners. SI improved 3.3-fold (TF-IDF proxy: 0.025 → 0.083). A
large-scale replication (\(n=30\)+) with unified measurement methodology
is needed to confirm these patterns.

\textbf{Ethics and scope.} Human participants in Exp1 provided informed
consent and participated voluntarily. The study involved minimal risk
(responding to hypothetical daily-life scenarios); all responses were
anonymized prior to analysis. Institutional review determinations for
this pilot are documented with the canonical dataset (Supplementary
§S11). Importantly, the core claims of this study---CKM's effect on LLM
consistency, the ablation decomposition, and cross-vendor behavioral
profiling---do not depend on the human pilot results. Exp1 provides
directional evidence that human participants can use the same F/H/E
categorization interface, but it is not part of the evidentiary basis
for Findings 1--5.

\hypertarget{state-persistence-and-practical-implications}{%
\subsubsection{7.4 State Persistence and Practical
Implications}\label{state-persistence-and-practical-implications}}

Turning from mechanism to application, state persistence response varied
dramatically across models (DFR: 0.01--0.85), reflecting fundamentally
different strategies for handling injected state. Low-DFR models treated
injected state as authoritative context, while high-DFR models treated
it as new information triggering re-evaluation. SDR of approximately 1.0
revealed that this re-evaluation preserved semantic content while
changing decisions, a pattern we characterize as meaning preservation
with decision re-evaluation. This distinction has direct practical
implications.

Applications requiring state consistency (e.g., multi-turn advisory
systems, persistent agent workflows) should select models based on DFR
profiles rather than capability scores alone, since a model with high
benchmark accuracy but high DFR may produce contradictory advice across
turns.

The generational improvement (6/12 primary NewModels achieving DFR = 0
vs.~0/19 Legacy) suggests that newer models are increasingly calibrated
for state persistence, though gemini-2.5-pro remains an outlier
(DFR\_persist = 0.644). More broadly, the DFR profile constitutes a
model fingerprint that could serve as a pre-deployment consistency
audit: organizations could run a standardized battery of
state-persistence trials to characterize each candidate model's
behavioral profile before committing to production deployment. Adaptive
state gating, in which injection strength is modulated based on a
model's measured DFR, represents a natural extension of this profiling
approach.

\hypertarget{limitations-1}{%
\subsubsection{7.5 Limitations}\label{limitations-1}}

The claims above are bounded by nine scope restrictions.

\begin{enumerate}
\def\labelenumi{\arabic{enumi}.}
\item
  \textbf{Human pilot size.} Exp1 used only six participants, so the
  human findings should be treated as directional rather than
  confirmatory.
\item
  \textbf{Consistency is not quality.} Higher SI does not show that the
  model reasons better; it only shows that responses converge more
  strongly under the tested protocol.
\item
  \textbf{Language scope.} All scenarios were presented in Korean.
  Cross-lingual validation is required before generalizing to other
  languages.
\item
  \textbf{Scenario scope.} The scenarios focused on student daily-life
  decisions. They do not yet test professional, safety-critical, or
  evolving real-world constraints.
\item
  \textbf{Embedding dependence.} SI and SDR depend on the embedding
  backbone (§6.4), although sensitivity analyses reduce concern about a
  single-backbone artifact.
\item
  \textbf{Temperature scope.} The temperature robustness probe assessed
  only \(T=0.7\).
\item
  \textbf{Power and substrate choice.} The Legacy parsed-field effect is
  well powered (\(g=1.18\), \(n=19\)), but the more conservative
  raw-output basis is less powered (\(g=0.61\), estimated post-hoc power
  \(\approx\) 0.55). The combined random-effects estimate
  (\(g_{\text{RE}}\) = 1.09, 31 model-level pairs) provides adequate
  power while preserving the high between-cohort heterogeneity (\(I^2\)
  = 77.6\%) as an explicit qualifier.
\item
  \textbf{Influential model.} gemini-2.5-pro shows persistently high DFR
  (0.644 in NewModels, 0.80 in Legacy) and influences Exp3 aggregate
  statistics. Sensitivity analyses excluding this model are reported in
  Supplementary §S-Exp3.
\item
  \textbf{Excluded deprecated model.} gpt-4.1-mini was excluded from
  primary analyses because OpenAI issued a deprecation notice during
  data collection. The case study (§6.6) and sensitivity analysis
  (Supplementary §S-Sensitivity) show that all directional findings are
  preserved under the 13-model cohort, but the primary NewModels sample
  is therefore \(n=12.\)
\end{enumerate}

\hypertarget{future-work}{%
\subsubsection{7.6 Future Work}\label{future-work}}

The limitations above define the next validation agenda. Nine directions
are especially important.

\begin{enumerate}
\def\labelenumi{\arabic{enumi}.}
\item
  \textbf{Mixed-effects variance decomposition.} Models such as SI
  \textasciitilde{} arm + (1\textbar model) + (1\textbar scenario) would
  clarify how much of the CKM effect is attributable to model-level
  versus scenario-level variation.
\item
  \textbf{Cross-lingual validation.} Matched scenarios across English,
  Korean, and Japanese would test whether the consistency gains
  generalize beyond Korean prompts.
\item
  \textbf{Decision-quality assessment.} Expert panels should evaluate
  whether reduced variability corresponds to better decisions, worse
  decisions, or merely more stable decisions.
\item
  \textbf{Formal state-trajectory analysis.} The chain data (§6.2)
  provide candidate measures for studying convergence and stability
  properties of CKM state trajectories.
\item
  \textbf{Adaptive state gating.} Deployment systems could adjust
  injection strength based on each model's measured DFR profile.
\item
  \textbf{Domain transfer.} Translation, code generation, medical
  reasoning, and professional decision tasks should be tested to
  determine whether CKM generalizes beyond student daily-life scenarios.
\item
  \textbf{Testimony axis extension.} Epistemological accounts often
  treat testimony as an independent source, distinct from perception and
  inference {[}56, 57{]}. Adding a fourth axis that separately tags
  communicated or attested information is a theoretically motivated
  extension requiring independent validation, and is left as future
  work.
\item
  \textbf{Extension beyond language models.} Robotic and physical AI
  systems combine sensor data, probabilistic inference, and priority
  signals. Testing whether F/H/E-like separation helps these systems
  would assess the broader generality of epistemic provenance
  enforcement.
\item
  \textbf{Standardized professional examinations.} Bar examinations,
  medical licensing exams, and certified public accountant tests would
  provide researcher-bias-free question pools, reproducible difficulty
  metadata, and domain variation in fact-inference ratios.
\end{enumerate}

The sham-restriction ablation previously identified as a future
direction has now been completed and is reported in §6.5. Its results
support the organization hypothesis over the compression hypothesis.

\begin{center}\rule{0.5\linewidth}{0.5pt}\end{center}

\hypertarget{conclusion}{%
\subsection{8. Conclusion}\label{conclusion}}

This paper tested a direct and bounded claim: \textbf{LLM behavioral
consistency can be measured, and part of the observed instability can be
reduced by forcing the model to separate facts, inferred assumptions,
and evaluative signals before deciding.} This is a claim about
behavioral stability under repeated and state-persistent trials, not a
claim about improved answer quality.

Across four core experiments, a 4-arm ablation, a 5-arm sham-restriction
ablation, and a temperature robustness probe, we evaluated 26 unique LLM
models from four vendors and two generations, producing 37,403
observations (35,475 primary + 1,928 case-study observations). The main
findings are:

\begin{enumerate}
\def\labelenumi{\arabic{enumi}.}
\item
  \textbf{CKM reduces repeated-output variability.} In the
  Korean-language student-decision scenarios tested here, F/H/E
  decomposition increased SI across both model generations: Legacy
  models showed 17/19 positive effects (parsed-field \(g=1.18\)),
  NewModels showed 12/12 positive effects (\(g=3.08\)), and the combined
  random-effects estimate was \(g_{\text{RE}}\) = 1.09 {[}0.83, 1.35{]}
  across 31 model-level pairs.
\item
  \textbf{The active mechanism is not JSON formatting alone.} The 4-arm
  ablation showed that F/H/E semantic separation survives value-only
  artifact control (\(g=2.24\)), whereas the apparent JSON-only effect
  disappears after structural syntax is removed. The critical
  intervention is therefore epistemic role separation: making the model
  distinguish what is observed, what is inferred, and what is
  evaluative.
\item
  \textbf{State persistence reduces decision flipping in newer models
  but remains model-dependent.} In the NewModels cohort, DFR dropped by
  82\% under CKM state persistence (0.385 \(\rightarrow\) 0.069,
  \(g=1.52\)), with 6/12 primary models reaching DFR = 0. However, DFR
  still varied widely across models, indicating that state-persistence
  compatibility is a measurable model property rather than a guaranteed
  feature of modern LLMs.
\item
  \textbf{Multi-turn instability is mainly a state-construction problem
  in this setting.} FCS was approximately 1.0 under fixed anchor states,
  suggesting that repeated stochastic sampling alone does not explain
  the observed multi-turn instability. The more important issue is how
  models construct, accept, and reuse structured state.
\item
  \textbf{Behavioral profiles reveal differences that capability
  benchmarks may miss.} Some models are stable in single-turn repetition
  but unstable when prior state is injected. This SI--DFR dissociation
  shows why consistency profiling should not be reduced to one score and
  why deployment audits may need repeated and state-persistent tests.
\item
  \textbf{CKM's effect decomposes into structural and reasoning-semantic
  components.} The 5-arm sham-restriction ablation estimated that
  structural scaffolding accounts for roughly 45\% of the SI
  improvement, while F/H/E reasoning-semantic content accounts for
  roughly 55\%. Only CKM achieved the joint pattern of SI improvement
  and DFR reduction.
\end{enumerate}

These results are intentionally bounded. They do not prove that CKM
makes LLMs more correct, more ethical, or more capable. They show that
behavioral consistency is an empirical property that can be profiled
across models and partially stabilized through prompt-level epistemic
state enforcement. Within the tested envelope---Korean-language student
daily-life decision scenarios, repeated-trial evaluation, and a 12-model
active NewModels cohort with Legacy cross-checks---CKM provides a
reproducible framework for measuring and reducing one form of LLM
behavioral instability.

The broader implication is that future LLM evaluation should ask not
only whether a model can produce a good answer once, but whether it can
maintain a stable reasoning state when the same decision problem recurs.
CKM is one candidate structure for that evaluation agenda. Further
validation across languages, domains, professional decision tasks, and
independent quality assessments is required before stronger claims can
be made.

\begin{center}\rule{0.5\linewidth}{0.5pt}\end{center}

\hypertarget{reproducibility-statement}{%
\subsection{Reproducibility Statement}\label{reproducibility-statement}}

All experiments used publicly available LLM APIs (OpenAI, Anthropic,
Google, xAI). API parameters: temperature = 0.7 (gpt-5-mini: \(T=1.0\),
the minimum supported by the OpenAI API for this model at collection
time); Study A pilot at \(T=0.\) Top-\(p\) left at provider defaults;
OpenAI \passthrough{\lstinline!seed!} parameter not used (no equivalent
on Anthropic / Google / xAI). Data collection: Legacy cohort 2026-02-15 to 2026-03-11; NewModels cohort 2026-03-10 to 2026-03-16
(Exp1--Exp3X, Study A/B); Exp6 sham arms 2026-04-23 to 2026-04-24. See Supplementary §S22.1 for the per-cohort timeline.

Per-vendor model IDs and the per-call snapshot string (where exposed by
the vendor) are documented in Supplementary §S22 and §S-App; readers should note
that vendor-side endpoint updates within the collection window may
produce non-identical reproductions even with matched IDs. Embedding
model:
\passthrough{\lstinline!sentence-transformers/paraphrase-multilingual-mpnet-base-v2!}
(768-dim) loaded via the \passthrough{\lstinline!sentence-transformers!}
library (release-pinned in the repository). Statistical seeds:
permutation tests use 10,000 resamples with seed = 42; bootstrap CIs use
2,000 resamples with seed = 42. Canonical analysis JSONs (e.g.,
\passthrough{\lstinline!ablation\_newmodels\_metrics\_v1.2\_active12.json!},
\passthrough{\lstinline!exp3extra\_newmodels\_metrics\_v1.2\_active12.json!},
\passthrough{\lstinline!exp6\_5way\_comparison\_v1.2\_active12.json!},
\passthrough{\lstinline!exp3\_newmodels\_metrics\_v1.3\_active12.json!})
are versioned in \passthrough{\lstinline!\_CANONICAL/!} directories with
SHA-256 checksums recorded in Supplementary §S2 and the repository
README.

The main text is synchronized against the accompanying Supplementary
Materials for section references, canonical dataset labels, and
claim-boundary wording. The primary numerical claims in the main text
follow the active-12 basis stated in §5.1 and the sensitivity reporting
in Supplementary §S-Sensitivity.

CKM prompt template, scenario texts (S1--S6, S9, S10), 4-arm
and 5-arm condition specifications, metric computation code, and
aggregated data are available at
https://github.com/TeenyToolSoftware/cogos-behavioral-consistency. The
canonical data state is the \passthrough{\lstinline!arxiv-v1.0!}
repository snapshot; no
canonical datasets or reported statistics have been altered for this
arXiv preparation.

\begin{center}\rule{0.5\linewidth}{0.5pt}\end{center}

\hypertarget{acknowledgments}{%
\subsection{Acknowledgments}\label{acknowledgments}}

We thank Youngjoon Suh (University of Illinois Chicago) for early
discussions, methodological feedback, and review comments during the
development of this work. We also thank the anonymous reviewers and the
automated LLM-based multi-agent audit pipeline (HAWK multi-pass internal
audit, three-vendor LLM cross-evaluation) that contributed to
claim-boundary tightening and readability improvements. Any remaining
errors are our own. This research received no specific grant from any
funding agency in the public, commercial, or not-for-profit sectors.

\begin{center}\rule{0.5\linewidth}{0.5pt}\end{center}

\hypertarget{references}{%
\subsection{References}\label{references}}

{[}1{]} Wei, J., et al.~(2022). Chain-of-Thought Prompting Elicits
Reasoning in Large Language Models. \emph{NeurIPS}.

{[}2{]} Yao, S., et al.~(2023). Tree of Thoughts: Deliberate Problem
Solving with Large Language Models. \emph{NeurIPS}.

{[}3{]} Wang, X., et al.~(2023). Self-Consistency Improves Chain of
Thought Reasoning in Language Models. \emph{ICLR}.

{[}4{]} Silberschatz, A., Galvin, P. B., \& Gagne, G. (2018).
\emph{Operating System Concepts} (10th ed.). Wiley.

{[}5{]} Significant Gravitas. (2023). AutoGPT. GitHub.

{[}6{]} Chase, H. (2022). LangChain. GitHub.

{[}7{]} Yao, S., et al.~(2023). ReAct: Synergizing Reasoning and Acting
in Language Models. \emph{ICLR}.

{[}8{]} Kahneman, D. (2011). \emph{Thinking, Fast and Slow}. Farrar,
Straus and Giroux.

{[}9{]} Flavell, J. H. (1979). Metacognition and Cognitive Monitoring.
\emph{American Psychologist}, 34(10), 906--911.

{[}10{]} Wood, D., Bruner, J. S., \& Ross, G. (1976). The Role of
Tutoring in Problem Solving. \emph{Journal of Child Psychology and
Psychiatry}, 17(2), 89--100.

{[}11{]} Besta, M., et al.~(2024). Graph of Thoughts: Solving Elaborate
Problems with Large Language Models. \emph{AAAI}.

{[}12{]} OpenAI. (2024). Structured Outputs in the API. Technical
Documentation.

{[}13{]} OpenAI. (2023). Function Calling and Other API Updates.
Technical Documentation.

{[}14{]} Madaan, A., et al.~(2023). Self-Refine: Iterative Refinement
with Self-Feedback. \emph{NeurIPS}.

{[}15{]} Shinn, N., et al.~(2023). Reflexion: Language Agents with
Verbal Reinforcement Learning. \emph{NeurIPS}.

{[}16{]} Laird, J. E. (2012). \emph{The Soar Cognitive Architecture}.
MIT Press.

{[}17{]} Anderson, J. R., et al.~(2004). An Integrated Theory of the
Mind. \emph{Psychological Review}, 111(4), 1036--1060.

{[}18{]} Card, S. K., Moran, T. P., \& Newell, A. (1983). \emph{The
Psychology of Human-Computer Interaction}. Lawrence Erlbaum.

{[}19{]} Bansal, G., et al.~(2021). Does the Whole Exceed its Parts? The
Effect of AI Explanations on Complementary Team Performance. \emph{CHI}.

{[}20{]} Steyvers, M., et al.~(2022). Bayesian Modeling of Human-AI
Complementarity. \emph{PNAS}, 119(11).

{[}21{]} Lai, V., \& Tan, C. (2019). On Human Predictions with
Explanations and Predictions of Machine Learning Models. \emph{FAccT}.

{[}22{]} Hu, E. J., et al.~(2022). LoRA: Low-Rank Adaptation of Large
Language Models. \emph{ICLR}.

{[}23{]} Rao, A. S., \& Georgeff, M. P. (1995). BDI Agents: From Theory
to Practice. \emph{ICMAS}.

{[}24{]} Reimers, N., \& Gurevych, I. (2019). Sentence-BERT: Sentence
Embeddings using Siamese BERT-Networks. \emph{EMNLP}.

{[}25{]} Park, J. S., et al.~(2023). Generative Agents: Interactive
Simulacra of Human Behavior. \emph{UIST}.

{[}26{]} Brown, T. B., et al.~(2020). Language Models are Few-Shot
Learners. \emph{NeurIPS}.

{[}27{]} Gigerenzer, G., \& Gaissmaier, W. (2011). Heuristic Decision
Making. \emph{Annual Review of Psychology}, 62, 451--482.

{[}28{]} Slovic, P., et al.~(2007). The Affect Heuristic. \emph{European
Journal of Operational Research}, 177(3), 1333--1352.

{[}29{]} Kadavath, S., et al.~(2022). Language Models (Mostly) Know What
They Know. \emph{arXiv:2207.05221}.

{[}30{]} Sclar, M., et al.~(2024). Quantifying Language Models'
Sensitivity to Spurious Features in Prompt Design. \emph{ICLR}.

{[}31{]} Zheng, C., et al.~(2024). Large Language Models Are Not Robust
Multiple Choice Selectors. \emph{ICLR}.

{[}32{]} Chen, W., et al.~(2023). Program of Thoughts Prompting.
\emph{TMLR}.

{[}33{]} Shyr, C., et al.~(2025). A Statistical Framework for Evaluating
the Repeatability and Reproducibility of Large Language Models.
\emph{medRxiv}.

{[}34{]} Wang, J., \& Wang, V. X. (2025). Assessing Consistency and
Reproducibility in the Outputs of Large Language Models.
\emph{arXiv:2503.16974}.

{[}35{]} Kim, M. H. (2025). Structured Cognitive Loop with a Governance
Layer. \emph{arXiv:2511.17673}.

{[}36{]} Kargupta, P., et al.~(2025). Cognitive Foundations for
Reasoning and Their Manifestation in LLMs. \emph{arXiv:2511.16660}.

{[}37{]} Wu, S., et al.~(2025). Cognitive LLMs: Toward Human-Like AI by
Integrating Cognitive Architectures and LLMs for Manufacturing
Decision-Making. \emph{SAGE Journals}.

{[}38{]} Xu, W., et al.~(2025). A-MEM: Agentic Memory for LLM Agents.
\emph{arXiv:2502.12110}.

{[}39{]} Sarin, S., et al.~(2025). Memoria: A Scalable Agentic Memory
Framework for Personalized Conversational AI. \emph{arXiv:2512.12686}.

{[}40{]} Chhikara, P., et al.~(2025). Mem0: Building Production-Ready AI
Agents with Scalable Long-Term Memory.

{[}41{]} Anthropic. (2025). Effective Context Engineering for AI Agents.
\emph{Anthropic Engineering Blog}.

{[}42{]} Herrera-Poyatos, A., et al.~(2025). An Overview of Model
Uncertainty and Variability in LLM-Based Sentiment Analysis.
\emph{Frontiers in AI}.

{[}43{]} Kang, J., Ji, M., Zhao, Z., \& Bai, T. (2025). Memory OS of AI
Agent. \emph{EMNLP 2025 (Oral)}. arXiv:2506.06326.

{[}44{]} Sumers, T. R., Yao, S., Narasimhan, K., \& Griffiths, T. L.
(2023). Cognitive Architectures for Language Agents.
\emph{arXiv:2309.02427}.

{[}45{]} Wu, J., et al.~(2025). Git Context Controller: Manage the
Context of LLM-based Agents like Git. \emph{arXiv:2508.00031}.

{[}46{]} Zhang, Y. \& Martinez, I. (2025). From Stochasticity to Signal:
A Bayesian Latent State Model for Reliable Measurement with LLMs.
\emph{arXiv:2510.23874}.

{[}47{]} Audi, R. (2011). \emph{Epistemology: A Contemporary
Introduction to the Theory of Knowledge} (3rd ed.). Routledge.

{[}48{]} BonJour, L. (1998). \emph{In Defense of Pure Reason}. Cambridge
University Press.

{[}49{]} Damasio, A. R. (1994). \emph{Descartes' Error: Emotion, Reason,
and the Human Brain}. Putnam.

{[}50{]} Tappolet, C. (2016). \emph{Emotions, Values, and Agency}.
Oxford University Press.

{[}51{]} de Sousa, R. (1987). \emph{The Rationality of Emotion}. MIT
Press.

{[}52{]} Matilal, B. K. (1986). \emph{Perception: An Essay on Classical
Indian Theories of Knowledge}. Clarendon Press.

{[}53{]} Crick, N. R. \& Dodge, K. A. (1994). A Review and Reformulation
of Social Information-Processing Mechanisms in Children's Social
Adjustment. \emph{Psychological Bulletin}, 115(1), 74--101.

{[}54{]} Lemerise, E. A. \& Arsenio, W. F. (2000). An Integrated Model
of Emotion Processes and Cognition in Social Information Processing.
\emph{Child Development}, 71(1), 107--118.

{[}55{]} Pessoa, L. (2008). On the Relationship between Emotion and
Cognition. \emph{Nature Reviews Neuroscience}, 9(2), 148--158.

{[}56{]} Coady, C. A. J. (1992). \emph{Testimony: A Philosophical
Study}. Clarendon Press.

{[}57{]} Burge, T. (1993). Content Preservation. \emph{The Philosophical
Review}, 102(4), 457--488.

{[}58{]} Hilgard, E. R. (1980). The Trilogy of Mind: Cognition,
Affection, and Conation. \emph{Journal of the History of the Behavioral
Sciences}, 16(2), 107--117.

{[}59{]} Clark, A. (2013). Whatever Next? Predictive Brains, Situated
Agents, and the Future of Cognitive Science. \emph{Behavioral and Brain
Sciences}, 36(3), 181--204.

{[}60{]} Stanovich, K. E. (2009). Distinguishing the Reflective,
Algorithmic, and Autonomous Minds: Is It Time for a Tri-Process Theory?
In J. S. B. T. Evans \& K. Frankish (Eds.), \emph{In Two Minds: Dual
Processes and Beyond} (pp.~55--88). Oxford University Press.

{[}61{]} Atil, I., Mitra, S., \& Viswanathan, K. (2025). Consistency in
Large Language Models. \emph{arXiv:2408.04667}.

{[}62{]} Haase, T., et al.~(2026). Understanding Within-LLM Variance: A
Variance Decomposition Approach. \emph{arXiv:2601.21339}.

{[}63{]} Du, Y., et al.~(2025). Flip Rates in Value Reasoning with Large
Language Models. \emph{EMNLP}. ACL Anthology: 2025.emnlp-main.395.

{[}64{]} Li, Z., et al.~(2025). A Survey on LLM Output Consistency:
Challenges and Future Directions. \emph{arXiv:2505.00268}.

{[}65{]} Xie, T., et al.~(2025). SCORE: Systematic COnsistency and
Robustness Evaluation of Large Language Models. \emph{NAACL Industry}.
arXiv:2503.00137.

{[}66{]} Zhang, Q. \& Zhu, K. (2025). Firm or Fickle? Evaluating the
Consistency of Large Language Models. \emph{ACL Findings}.
arXiv:2503.22353.

{[}67{]} Krishnamurthy, S., et al.~(2026). NabaOS: An Operating System
for Artificial Intelligence Agents Based on Indian Epistemology.
\emph{arXiv:2603.10060}.

{[}68{]} Manson, R. (2025). MOLES: Modelling Epistemic Stances of Large
Language Models. Preprint.

{[}69{]} Zhang, Y., et al.~(2024). Cognitive Kernel: An Open-Source
Agent System Towards Generalist Autopilot. \emph{NAACL 2025 Demo}.
arXiv:2409.10277.

{[}70{]} Pei, K., et al.~(2025). Behavioral Fingerprinting of Large
Language Models. \emph{arXiv:2509.04504}.

{[}71{]} Laban, P., et al.~(2024). Are You Sure? Grounding LLM Factual
Accuracy Under Adversarial Challenge. \emph{EMNLP}.

{[}72{]} Li, R., et al.~(2025). Measuring and Improving Attentional
Sycophancy in Large Language Models. \emph{ACL Findings}.

{[}73{]} Lee, S., et al.~(2025). Confirmation Bias in LLM
Decision-Making Under Counter-Evidence. Preprint.

{[}74{]} Guan, X., et al.~(2026). Evaluating Action Consistency in GUI
Agents. Preprint.

{[}75{]} Huang, J.-t., et al.~(2025). On the Failure of Latent State
Persistence in Large Language Models. \emph{arXiv:2505.10571} (preprint;
accessed 2026-05-22).

{[}76{]} Tosato, T., et al.~(2025). Persistent Instability in LLM's
Personality Measurements: Effects of Scale, Reasoning, and Conversation
History (PERSIST). \emph{arXiv:2508.04826} (preprint; accessed
2026-05-22).

{[}77{]} Leshin, J., Shah, M., Timmis, I., \& Kang, D. (2026).
Behavioral Fingerprints for LLM Endpoint Stability and Identity.
\emph{arXiv:2603.19022} (preprint; accessed 2026-05-22).

{[}78{]} Luo, H., \& Laban, G. (2026). SPASM: Stable Persona-driven
Agent Simulation for Multi-turn Dialogue Generation.
\emph{arXiv:2604.09212} (preprint; accessed 2026-05-22).

{[}79{]} Liu, Y., Zhu, M., Liu, S., Hu, B., \& Zhang, L. (2026).
Enhancing Persona Following at Decoding Time via Dynamic Importance
Estimation for Role-Playing Agents. \emph{arXiv:2603.01438} (preprint;
accessed 2026-05-22).

{[}80{]} Kim, M., Im, S., Choi, J., Lee, J., Shim, C., Hong, H., \&
Choi, E. (2026). PICon: A Multi-Turn Interrogation Framework for
Evaluating Persona Agent Consistency. \emph{arXiv:2603.25620} (preprint;
accessed 2026-05-22).

{[}81{]} Cavalin, P., Sanctos, C., Grave, M., Pinhanez, C., \&
Primerano, Y. (2025). CAT: A Metric-Driven Framework for Analyzing the
Consistency-Accuracy Relation of LLMs under Controlled Input Variations.
\emph{arXiv:2512.23711} (preprint; accessed 2026-05-22).

{[}82{]} Hsing, N. (2025). MIRROR: Modular Internal Processing for
Personalized Safety in Multi-turn Dialogue. \emph{OpenReview lwSV507BPm}
(ICLR 2026 withdrawn submission, 21 Nov 2025; cited as prior art under
withdrawal status).

\end{document}